\documentclass[journal]{IEEEtran}

\ifCLASSINFOpdf
\else
\fi

\usepackage{hyperref}       
\usepackage{url}            
\usepackage{booktabs}       
\usepackage{nicefrac}       
\usepackage{microtype}      
\usepackage{dsfont}
\usepackage{subcaption}
\usepackage{graphicx}
\usepackage{xcolor}
\usepackage{url}
\usepackage{amsmath, amsthm, amsfonts, amssymb}
\usepackage{bbm}
\usepackage{newtxtext,newtxmath}
\usepackage[ruled,vlined]{algorithm2e}
\usepackage{cite}

\usepackage{array, makecell}
\usepackage{multirow}
\usepackage{arydshln}
\usepackage[flushleft]{threeparttable}
\usepackage[export]{adjustbox}

\begin{document}

\title{Early Detection of COVID-19 Hotspots Using Spatio-Temporal Data}
%
%
%

\author{Shixiang~Zhu, 
        Alexander~Bukharin,
        Liyan~Xie,
        Khurram~Yamin,
        Shihao~Yang,
        Pinar~Keskinocak,
        and~Yao~Xie 
\thanks{S. Zhu, A. Bukharin, L. Xie, S. Yang, P. Keskinocak, and Y. Xie was with the H. Milton Stewart School of Industrial and Systems Engineering, Georgia Institute of Technology, Atlanta,
GA, 30332 USA. e-mail: yao.xie@isye.gatech.edu.}
}

\maketitle

\begin{abstract}
Recently, the Centers for Disease Control and Prevention (CDC) has worked with other federal agencies to identify counties with increasing coronavirus disease 2019 (COVID-19) incidence (hotspots) and offers support to local health departments to limit the spread of the disease.
Understanding the spatio-temporal dynamics of hotspot events is of great importance to support policy decisions and prevent large-scale outbreaks.
This paper presents a spatio-temporal Bayesian framework for early detection of COVID-19 hotspots (at the county level) in the United States. 
We assume both the observed number of cases and hotspots depend on a class of latent random variables, which encode the underlying spatio-temporal dynamics of the transmission of COVID-19. 
Such latent variables follow a zero-mean Gaussian process, whose covariance is specified by a non-stationary kernel function.
The most salient feature of our kernel function is that deep neural networks are introduced to enhance the model's representative power while still enjoying the interpretability of the kernel. 
We derive a sparse model and fit the model using a variational learning strategy to circumvent the computational intractability for large data sets. Our model demonstrates better interpretability and superior hotspot-detection performance compared to other baseline methods. 
\end{abstract}

\begin{IEEEkeywords}
COVID-19 hotspots, Gaussian processes, non-stationary kernel, spatio-temporal model.
\end{IEEEkeywords}

%
\IEEEpeerreviewmaketitle

\section{Introduction}

\IEEEPARstart{T}{he} ongoing global pandemic caused by the coronavirus disease (COVID-19) has spread rapidly over more than 200 countries in the world since its emergence in 2019. 
Even the largest economies' resources have been strained due to the spread of COVID-19. 
Predicting potential hotspots ahead of time can play a significant role in deploying targeted interventions, such as testing, tracing, and isolation, and slow down the disease spread \cite{oster2020trends}. 

Large-scale, population-based testing can indicate regional hotspots, but at the cost of a delay between testing and actionable results. 
Accurately identifying changes in the infection rate requires sufficient testing coverage of a given population, which can be costly and requires substantial testing capacity.
Regional variation in testing access can also hamper the ability of public health organizations to detect rapid changes in infection rates.
Recent studies \cite{CDC} aimed at estimating the spread of COVID-19 by forecasting the number of confirmed cases or the number of deaths. 
However, these methods failed to provide a satisfactory case prediction accuracy.
Therefore, there is a high unmet need for tools and methods that can facilitate the timely and accurate identification of infection hotspots and enable policymakers to act effectively with minimal delay \cite{rossman2020framework}.

\begin{figure}[!t]
\centering
\begin{subfigure}[h]{0.49\linewidth}
\includegraphics[width=\linewidth]{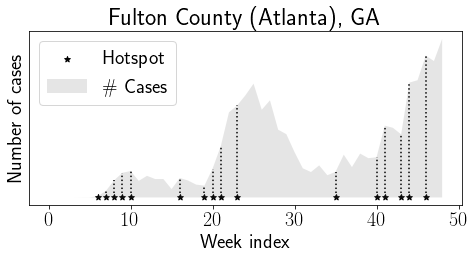}
\end{subfigure}
\begin{subfigure}[h]{0.49\linewidth}
\includegraphics[width=\linewidth]{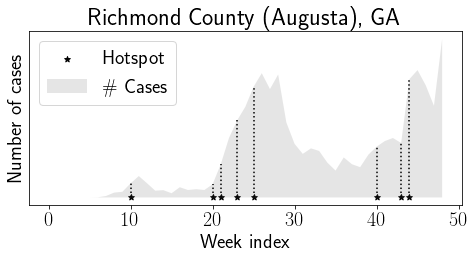}
\end{subfigure}
\begin{subfigure}[h]{0.49\linewidth}
\includegraphics[width=\linewidth]{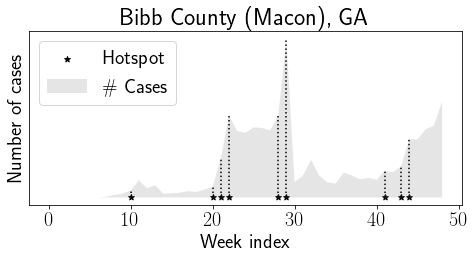}
\end{subfigure}
\begin{subfigure}[h]{0.49\linewidth}
\includegraphics[width=\linewidth]{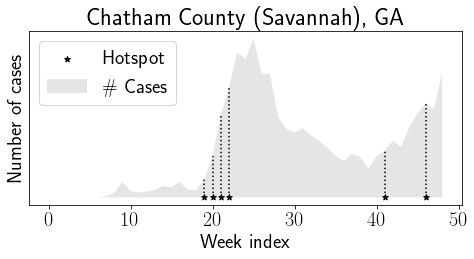}
\end{subfigure}
\caption{Examples of COVID-19 hotspots identified by CDC \cite{oster2020trends}. In general, these hotspots define the onset of a local outbreak of COVID-19.}
\label{fig:illustrative-example}
\end{figure}

The Centers for Disease Control and Prevention (CDC) with other federal agencies have identify counties with a significant increase in COVID-19 incidence (hotspots) \cite{oster2020trends}, which offers a unique opportunity to investigate the spatio-temporal dynamics between the identified hotspots.
Fig.~\ref{fig:illustrative-example} gives some real examples of the hotspots at four different counties in the state of Georgia. 
The identified hotspots indicate the relative temporal increases in confirmed cases and mark the onset of local outbreaks.

In this paper, we propose an effective COVID-19 hotspot detection framework that utilizes the hotspot data and multiple other data sources, including community mobility, to enhance hotspot detection accuracy. 
We assume the hotspot and number of cases in the same location depending on common priori factors, represented by a latent spatio-temporal random variable.
This latent variable is modeled by a Gaussian process, whose covariance is characterized by an interpretable non-stationary kernel.
We note that the non-stationarity of our kernel plays a pivotal role in the success of our model because the spread of the virus shows heterogeneous spatial correlation across different regions. 
For example, the virus is likely to spread more slowly in a sparsely populated area such as rural Nebraska compared to a densely populated area such as New York City. 
We formulate our kernel function using carefully crafted feature functions incorporating neural networks, which provide greater flexibility in capturing the complex dynamics of the spread of COVID-19 while still being highly interpretable. 
To tackle the computational challenge of the Gaussian process with a large-scale data set, we also derive a sparse model and fit the model efficiently via a variational learning strategy. 

The remainder of the paper is organized as follows. We first discuss the literature relevant to COVID-19 hotspot detection and other related work. We describe the data sets in Section~\ref{sec:data}. We introduce the proposed hotspot detection framework in  Section~\ref{sec:framework}. We present an efficient computation strategy and the learning algorithm for our detection framework in 
Section~\ref{sec:efficient-computation}. 
Finally, we present the interpretation of our model and the numerical results on COVID-19 data in Section~\ref{sec:results}.


\vspace{.1in}
\noindent\emph{Related work.}
Hotspot detection is closely related to the prediction of confirmed cases. Therefore, we first review some prediction methods for completeness. Compartmental models are mathematical modeling of infectious diseases and have been widely used in epidemiology. In simple SIR models, \cite{Harko2014}, the population is assigned to compartments with labels S (susceptible), I (infectious), and R (recovered), respectively. The transition rates between compartments are typically modeled using differential equations. 
Extensions and variants of SIR models include the SIRD model \cite{fernandez2020estimating,caccavo2020chinese} which considers deceased individuals, and the SEIR model \cite{Hethcote2000,yang2020modified,hou2020effectiveness} which considers exposed periods, to name a few. Compartmental models work well when applied to large regions/populations, such as a state or a country because they assume a fixed/closed population. 
However, populations between geographic areas, such as counties, may interact with each other in the desired high-resolution modeling. Therefore, we use a spatio-temporal model that is more flexible and can capture the spread between different counties. 

Much work has been done on predicting the number of COVID-19 cases and deaths at the national level or state level, without considering the spatial correlation across smaller regions \cite{LANL, UTaustin, MOBS, IHME, zhu2020high,bertozzi2020challenges,ghosh2020covid}. 
Machine learning-based approaches have also been considered in \cite{alazab2020covid}.
Some work \cite{tamang2020forecasting} attempts to use neural networks to model the accumulative number of confirmed cases.
Recurrent neural network-based methods \cite{hawas2020generated, zhao2020well} have been applied to model the temporal dynamics of the COVID-19 outbreak. Moreover, online COVID-19 forecasting tools include the COVID-19 simulator \cite{chhatwal2020covid} and the COVID-19 Policy Alliance developed by a group in MIT. In this paper, hotspot detection is a binary classification problem, which differs in nature from the regression investigated by these studies. 

Besides accurate prediction and detection, understanding the spatial spread underlying the COVID-19 outbreak is also of great importance. An interpretable spatial model can help the government develop efficient public health policies to slow the spread during the early stages of COVID-19.
Compared with literature that focuses on prediction, studies evaluating the spatial spread of the COVID-19 pandemic are still limited \cite{poirier2020real}. 
Previously, the spatial spread has been studied for the outbreak of severe acute respiratory syndrome (SARS) in Beijing, and mainland China \cite{Meng2005, Fang2009, Kang2020, Jia2020, CCDCP2020,Meng2005} using only limited or localized data. In \cite{chiang2020hawkes}, the multivariate Hawkes process has been applied to model the conditional intensity of new confirmed COVID-19 cases and deaths in the U.S. at the county-level, without considering the influence from the big cities (main transportation hubs) and other important demographic factors. 
In \cite{altieri2020curating}, two types of county-level predictive models are developed based on the exponential and linear model, respectively. It focuses on modeling the dynamics of cumulative death counts.
In \cite{kapoor2020examining}, graph neural networks are adopted to capture the spatio-temporal dynamics between various features; however, a common disadvantage of the neural network-based methods is the lack of interpretability, which hinders from further understanding the mechanism underlying the COVID-19 spread.

Few studies have so far been conducted to investigate COVID-19 hotspots and their early detection. 
Similar to the CDC's definition, a recent study \cite{varsavsky2021detecting} considers a sudden increase in the number of cases in a specific geographical region. Unlike hotspot detection, they focus on estimating disease prevalence using logistic regression based on both symptoms and swab test results. 
In \cite{shariati2020spatiotemporal}, hotspots are defined as spots with the highest incidence rate. This paper adopts statistical and spatial analysis to determine the spatial distribution and spatial clustering patterns of the COVID-19 incidence rate. To identify the COVID-19 hotspots, the Getis-Ord spatial statistic \cite{getis2010analysis} was then applied. 
In another work \cite{feng2021topological}, topological data analysis was applied to identify the hotspots of COVID-19 infections, which is defined as regions with higher case counts than their surrounding areas. However, no quantitative results, such as the estimation of confirmed cases or the prediction of future hotspots, were provided in \cite{feng2021topological}.

Many studies used the Gaussian process for COVID-19 case prediction.
In \cite{dhamodharavadhani2021covid}, a Gaussian process regression model is applied to mortality rate prediction in India. Unlike our model, \cite{dhamodharavadhani2021covid} does not consider any spatial factors in the spread of COVID-19 and instead predicts cases on a national level. Some recent work \cite{velasquez2020forecast, ketu2021enhanced} have used Gaussian process models with a squared exponential kernel to forecast cases in the United States. However, these works provide state and national-level forecasts less granular than the city- or county-level forecasts produced by our model. In addition, \cite{velasquez2020forecast, ketu2021enhanced} use a stationary kernel, meaning that the kernel cannot adapt to different spatial patterns in different locations like the non-stationary spatial kernel discussed in this paper.




\section{COVID-19 Data Description}
\label{sec:data}

The data sets we used in our study include the number of cases and deaths, COVID-19 hotspots identified by the Centers for Disease Control and Prevention (CDC), and community mobility provided by Google. The study period is from March 15, 2020, to January 17, 2021, consisting of 50 weeks and 3,144 US counties. We excluded the data after February 2021, when a large-scale COVID-19 vaccine rollout had been launched across the United States, which significantly shifted the dynamics of the COVID-19 spread.

\vspace{.05in}
\noindent\emph{Confirmed cases and deaths.}
We used the data set from The New York Times (NYT) \cite{NYT2019}\footnote{One reason we use the NYT data rather than Johns Hopkins (JHU) data \url{https://coronavirus.jhu.edu/map.html} is that JHU data have retrospective data revision (when state update the COVID-19 definition, or have data error, etc), while NYT data never revise its history.} which
includes two parts: (i) \emph{confirmed cases} are counts of individuals whose coronavirus infections were confirmed by a laboratory test and reported by a federal, state, territorial, or local government agency;
(ii) \emph{confirmed deaths} are individuals who have died and meet the definition for a confirmed COVID-19 case. 
In practice, we have observed periodic weekly oscillations in daily reported cases and deaths, which could have been caused by testing bias (higher testing rates on certain days of the week). To reduce such bias, 
we aggregate the number of cases and deaths of each county {\it by week}. 

\begin{figure}[!t]
\centering
\begin{subfigure}[h]{0.49\linewidth}
\includegraphics[width=\linewidth, frame]{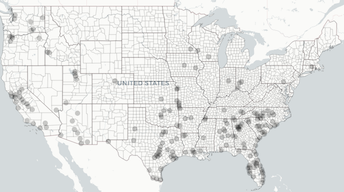}
\caption{June 28, 2020}
\end{subfigure}
\begin{subfigure}[h]{0.49\linewidth}
\includegraphics[width=\linewidth, frame]{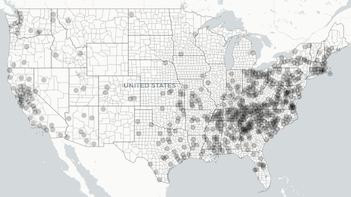}
\caption{December 20, 2020}
\end{subfigure}
\caption{Snapshots of hotspots identified by CDC. The black circles indicate the counties that have been identified as hotspots in that week.}
\label{fig:data-hotspot}
\end{figure}

\vspace{.05in}
\noindent\emph{Hotspots.}
On May 7, 2020, the CDC and other federal agencies began identifying counties with increasing COVID-19 hotspots to better understand transmission dynamics and offer targeted support to health departments in affected communities. The CDC identified hotspots daily starting on January 22, 2020, among counties in U.S. states and the District of Columbia by applying standardized criteria developed through a collaborative process involving multiple federal agencies \cite{oster2020trends, oster2020transmission}. In general, hotspots were defined based on relative temporal increases in the number of cases.
To match the temporal resolution with the number of cases and deaths, we expand the definition of a hotspot from daily-level to weekly-level. A week is identified as a hotspot if it contains at least one hotspot day identified by CDC.
The weekly number of counties meeting hotspot criteria peaked in early April, decreased and stabilized during mid-April–early June, then increased again during late June--early July. The percentage of counties in the South and West Census regions meeting hotspot criteria increased from 10\% and 13\%, respectively, during March--April to 28\% and 22\%, respectively, during June--July. Fig.~\ref{fig:data-hotspot} gives snapshots of the identified hotspots at two particular weeks. 

\begin{figure}[!ht]
\centering
\begin{subfigure}[h]{0.48\linewidth}
\includegraphics[width=\linewidth, frame]{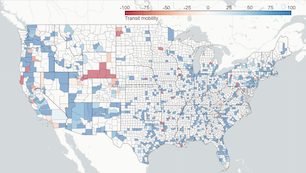}
\caption{Transit on March 1, 2020}
\end{subfigure}
\begin{subfigure}[h]{0.48\linewidth}
\includegraphics[width=\linewidth, frame]{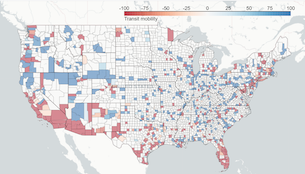}
\caption{Transit on July 12, 2020}
\end{subfigure}
\vfill
\begin{subfigure}[h]{0.48\linewidth}
\includegraphics[width=\linewidth, frame]{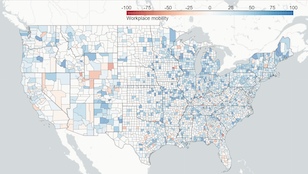}
\caption{Workplace on March 1, 2020}
\end{subfigure}
\begin{subfigure}[h]{0.48\linewidth}
\includegraphics[width=\linewidth, frame]{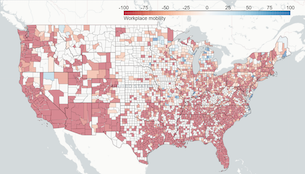}
\caption{Workplace on July 12, 2020}
\end{subfigure}
\caption{Overview of Google mobility data in two selected categories: workplace and transit on two different days. Counties in red and blue indicate their mobility is lower and higher than the normal level, respectively. The mobility level varies over time and space due to local government policy changes in response to COVID-19.}
\label{fig:data-mobility}
\end{figure}

\vspace{.05in}
\noindent\emph{Community mobility.}
The COVID-19 Community Mobility Reports \cite{Covidcommunity2020} 
record people's movement by county daily, across various categories such as retail and recreation, groceries and pharmacies, parks, transit stations, workplaces, and residential. The data shows how visitors to (or time spent in) categorized places change compared to the baseline days (in percentage). The negative percentage means that the level of mobility is lower than the baseline, and the positive percentage represents the opposite. The mobility on a baseline day represents a normal value for that day of the week. This mobility report sets the baseline as the median value from the five weeks from January 3rd to February 6th, 2020. 
Similar to the two data sets mentioned above, we aggregate each county's mobility data by week. 
Examples of two categories, transit stations, and workplaces, are shown in Fig.~\ref{fig:data-mobility}.

\section{COVID-19 Hotspot Detection Framework}
\label{sec:framework}

This section presents our hotspot detection framework, consisting of two spatio-temporal models: confirmed cases and hotspots.
Consider weeks $\mathscr{T} = \{t=1, \dots, T\}$ starting from March 15, 2020 to January 17, 2021 and locations (counties) $\mathscr{I} = \{i=1, \dots, I\}$, 
with latitude and longitude $s_i \in \mathscr{S} \subset \mathbb{R}^2$, $i \in \mathscr{I}$, where $\mathscr{S}$ represents the space of geographic coordinate system (GCS). 
The two models, respectively, focus on weekly confirmed cases $y_{it} \in \mathbb{Z}_+$ and identified hotspots $h_{it} \in \{0, 1\}$ of COVID-19 at location $i \in \mathscr{I}$ and time $t \in \mathscr{T}$, where $h_{it} = 1$ if there is a hotspot at location $i$ and time $t$, and 0, otherwise. 

CDC \cite{oster2020trends} defined the hotspots based on relative temporal increases in the number of cases, i.e., the occurrence of the hotspots depends on the spatio-temporal correlation across different locations and over time, and {\it not} on the mean number of cases (see the observation in Fig.~\ref{fig:illustrative-example}).
Hence, we capture the correlation between $y_{it} $ and  $h_{it}$ by connecting these two models in the spatio-temporal space $(t, s_i)$ through a latent spatio-temporal random variable $f(t, s)$, characterized by a Gaussian process (GP) with zero mean and covariance specified by a kernel function $k$.

The goal is to find the optimal pair of these two models that best predict the hotspots and the cases for one week ahead. We refer to the proposed framework as the spatio-temporal Gaussian process (\texttt{STGP}).

\subsection{Spatio-Temporal Gaussian Process (STGP) Models}
\label{sec:spatio-temporal-gaussian}

For the notational simplicity, we first
denote the spatio-temporal coordinate $(t, s_i)$ by $\mathbf x_{it} \in \mathscr{X}$, where $\mathscr{X} \coloneqq \mathscr{T} \times \mathscr{S} \subset \mathbb{R}^3$ represents the spatio-temporal space.
For any subset $\mathbf{X} \subseteq \mathscr{X}$ with $N$ spatio-temporal coordinates, the set of function variables $\mathbf{f} \coloneqq \{f(\mathbf{x})\}_{\mathbf{x}\in\mathbf{X}}$ has joint zero-mean Gaussian distribution
\begin{equation}
    p(\mathbf{f}) = \mathcal{N}(\mathbf{0}, \mathbf{K}_{XX}),
    \label{eq:prior}
\end{equation}
where $\mathbf{K}_{XX}$ is a $N \times N$ matrix and its entries are pairwise evaluations of $k(\mathbf{x}, \mathbf{x}')$, $\forall \mathbf{x}, \mathbf{x}' \in \mathbf{X}$.

\vspace{.1in}
\noindent\emph{Case model.}
We define a spatio-temporal model for the confirmed cases 
in the following form:
\begin{equation}
    y_{it} = \mu_{it} + f(\mathbf x_{it}) + \epsilon_{it},~i\in\mathscr{I},t\in\mathscr{T},
    \label{eq:y-conditional-f}
\end{equation}
where $\epsilon_{it} \sim \mathcal{N}(0, \sigma_\epsilon^2)$ is assumed to be i.i.d. normally distributed; $\mu_{it}$ is the mean of number of confirmed cases at time $t$ in location $i$. 
For a set of $N$ observed spatio-temporal coordinates $\mathbf{X}$, we denote the number of confirmed cases and their means as $\mathbf{y} \coloneqq \{y_{it}\}_{\mathbf{x}_{it}\in\mathbf{X}}$ and $\boldsymbol{\mu} \coloneqq \{\mu_{it}\}_{\mathbf{x}_{it}\in\mathbf{X}}$, respectively.

We assume the mean of the number of confirmed cases at a particular location relates to the covariates in its nearby counties according to an underlying undirected graph $\mathscr{G} = (\mathscr{I}, \mathscr{E})$, where $\mathscr{I}$ is the set of vertices representing all the locations, and $\mathscr{E} \subseteq \{(i, j) \in \mathscr{I}^2\} $ is a set of undirected edges representing the connections between locations.
There is an edge between two vertices whenever the corresponding locations are geographically adjacent.
Let 
$\boldsymbol{\eta}_{it} \coloneqq [\eta_{it1},\dots,\eta_{itl},\dots,\eta_{itL}]^\top \in \mathbb{R}^L$ denote the data of these covariates at location $i \in \mathscr{I}$ and time $t \in \mathscr{T}$, and let
$\boldsymbol{\omega}_{it} \coloneqq [\omega_{it1}, \dots, \omega_{itl}, \dots, \omega_{itL}]^\top \in \mathbb{R}^L$ denote the parameters of the corresponding covariates; 
$L$ denotes the number of features. 
In practice, we use the number of confirmed cases, the number of deaths, and six community mobilities variables in the past two weeks as the input covariates with $L = 16$. 
Formally, we define $\mu_{it}$ as 
\begin{equation}
    \mu_{it} = \sum_{\tau \in \mathcal{H}_t} \sum_{j:(i,j)\in \mathscr{E}} \boldsymbol \eta_{j \tau}^\top \boldsymbol\omega_{j\tau},~\forall i \in \mathscr{I}, t \in \mathscr{T},
    \label{eq:mean}
\end{equation}
where $\mathcal{H}_t = \{\tau: t-d \le \tau < t \}$ represent the recent history with memory depth $d < T$.


\vspace{.1in}
\noindent\emph{Hotspot model.}
We express the conditional probability of the hotspots $\mathbf{h} \coloneqq \{h_{it}\}_{\mathbf{x}_{it} \in \mathbf{X}}$ for a set of spatio-temporal coordinates $\mathbf{X}$ as:
\begin{equation}
    p(\mathbf{h}|\mathbf{f}) = \prod_{\mathbf{x}_{it} \in \mathbf{X}} \mathcal{B}(h_{it}|\phi(f(\mathbf{x}_{it}))),
    \label{eq:h-conditional-f}
\end{equation}
where $\mathcal{B}(h_{it}|\phi(f(\mathbf{x}_{it}))) = \phi(f(\mathbf{x}_{it}))^{h_{it}} (1-\phi(f(\mathbf{x}_{it})))^{1-h_{it}}$ is the likelihood for the Bernoulli distribution and $\phi$ is a sigmoid function. 

\vspace{.1in}
\noindent\emph{Learning objective.}
We aim to detect the hotspot while taking advantage of the information that has been recorded in the number of confirmed cases. To this end, we learn the model by optimizing the following combined objective:
\begin{equation}
    \underset{\boldsymbol \theta \in \Theta}{\max} ~\ell(\boldsymbol{\theta}) \coloneqq \ell_{h}({\boldsymbol \theta}) + \delta \ell_{y}(\boldsymbol \theta),
    \label{eq:objective}
\end{equation}
where $\delta>0$ controls the ratio between two objectives and $\boldsymbol{\theta} \in \Theta$ is the set of parameters defined in the kernel $k$. 
The $\ell_{y}({\boldsymbol \theta}) \coloneqq \log p(\mathbf{y})$ denotes the log marginal likelihood of observed confirmed cases and $\ell_{h}({\boldsymbol \theta}) \coloneqq \log p(\mathbf{h})$ denotes the log marginal likelihood of observed hotspots.
We note that log marginal likelihood of cases in the second term plays a key role in ``regularizing'' the model by leveraging the information in the case records as shown in Appendix~\ref{append:cv-exp-results}.
We also present the $5$-fold cross-validation that quantitatively measures the $F_1$ score of the hotspot detection and the mean square error of the case prediction with different $\delta$ in Fig.~\ref{fig:cv-delta-outofsample} (Appendix~\ref{append:cv-exp-results}). The result confirms that the appropriate choice of $\delta$ can significantly improve the performance of hotspot detection. 


\subsection{Spatio-Temporal Deep Neural Kernel}

We discuss the choice of the kernel function $k$ in this subsection. Standard GP models use a stationary covariance, in which the covariance between any two points is a function of their Euclidean distance. However, stationary GPs fail to adapt to variable smoothness in the function of interest. 
This is of particular importance in geophysical and other spatial data sets, in which domain knowledge suggests that the function may vary more quickly in some parts of the input space than in others.
For example, COVID-19 is likely to be spreading slower than in sparsely versus densely populated regions. Here, we consider the following non-stationary spatio-temporal kernel:
\begin{equation}
    k(t, t', s, s') = \nu(t, t') \cdot \left(\sum_{r=1}^R w^{(r)}_{s'} \upsilon^{(r)}(s, s')\right),
    \label{eq:spatio-temporal-kernel}
\end{equation}
where $\nu(t, t')$ is a stationary kernel that captures temporal correlation between time $t$ and $t'$; $\upsilon^{(r)}(s, s')$ is a component of the non-stationary spatial kernel which evolves over the space and $w^{(r)}_{s'}$ is the corresponding weight satisfying $\sum_{r=1}^R w^{(r)}_{s'} = 1$. $R$ is the number of components considered.
By likening the relationship between the spatial kernel component to that of the Gaussian component in the Gaussian mixture, we seek to enhance the representative power of our kernel by adding more independent components to the spatial kernel.

\vspace{.1in}
\noindent\emph{Stationary temporal kernel.}
We define the kernel function that characterizes the temporal correlation between $t,t' \in \mathscr{T}$ as an stationary Gaussian function: 
\[
\nu(t, t') = \exp\left\{- \frac{1}{2 \sigma_\nu^2} ||t-t'||^2\right\},
\]
where $\sigma_\nu \in \mathbb{R}_+$ is the bandwidth parameter. This kernel function hypothesizes that the virus' transmission is highly related to its recent history and their correlation will decay exponentially over time.

\vspace{.1in}
\noindent\emph{Non-stationary spatial kernel.}
To account for non-stationarity, we now allow the smoothing kernel to depend on spatial location $s$. 
For ease of discussion and simplicity of notation, we omit the superscript $r$ in $\upsilon^{(r)}(s, s')$ and $w^{(r)}_{s'}$, and present the structure of a single non-stationary spatial kernel component. 
We use $\kappa_s(\cdot)$ to denote a kernel which is centered at the point $s$ and whose shape is a function of location $s$. Once $\kappa_s(\cdot)$ is specified for all $s \in \mathscr{S} \subseteq \mathbb{R}^2$, the correlation between two points $s$ and $s'$ is then 
\begin{equation}
    \upsilon(s, s') \propto \int_{\mathbb{R}^2} \kappa_s(u) \kappa_{s'}(u) du. 
    \label{eq:def-non-stationary-kernel}
\end{equation}
Because of the constructive formulation under the moving average specification, the resulting correlation function $\upsilon(s, s')$ is certain to be positive definite.
We favor working with the kernels $\kappa_s(\cdot)$ rather than directly with the correlation function $\upsilon(s, s')$ since this makes it difficult to ensure positive symmetry definiteness for all $s$ and $s'$.
Following the idea of \cite{bernardo1998non, zhu2021imitation}, we define each $\kappa_s(\cdot)$ to be a normal kernel centered at $s$ with spatially varying covariance matrix $\Sigma_s$. In this case given the parameterized $\Sigma_s$ and $\Sigma_{s'}$,
the correlation function is given by an easy to compute formula
\[
\upsilon(s, s') \!\propto\! \frac{| \Sigma_s + \Sigma_{s'} |^{-\frac{1}{2}}}{2\pi} \exp\left \{-\frac{1}{2}(s' - s)^\top(\Sigma_s + \Sigma_{s'})^{-1}(s' - s)\right \}.
\]
The derivation of this formula can be found in Appendix~\ref{append:proof-non-stationary-kernel}.

To assure that the kernel $\{\kappa_s(\cdot)\}$ vary smoothly over space $\mathscr{S}$, we parameterize $\Sigma_s$ and then allow the parameters to evolve with location. 
For this paper we will focus on a geometrically based specification which readily extends beyond the use of the Gaussian kernel considered here.

\begin{figure}[!t]
\centering
\includegraphics[width=\linewidth]{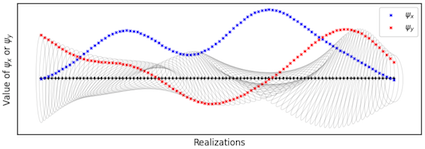}
\caption{An example of the randomly generated focus points $\boldsymbol{\psi}$ and their corresponding covariance $\Sigma$. The horizontal coordinate represents different focus points realizations; the vertical coordinate represents the value of these realizations. The ellipses portray the shape of the corresponding covariance for the kernel $\kappa_s(\cdot)$ associated with that location $s$.}
\label{fig:focus-points-exp}
\end{figure}

There is a one-to-one mapping from a bivariate normal distribution to its one standard
deviation ellipse, so we define a spatially varying family of ellipses which, in turn, defines the spatial distribution for $\Sigma_s$. 
Let the two focus points in $\Psi \subset \mathbb{R}^2$ denoted by $\boldsymbol \psi_s \coloneqq (\psi_x(s), \psi_y(s)) \in \Psi$ and $- \boldsymbol \psi_s \coloneqq (-\psi_x(s), -\psi_y(s)) \in \Psi$ define an ellipse centered at $s$ with fixed area $A$. 
This then corresponds to the Gaussian kernel with covariance matrix $\Sigma_s$ defined by 
\begin{equation}
\begin{aligned}
    \Sigma_s = \lambda^2 
    \begin{pmatrix} 
    Q + \frac{\|\boldsymbol\psi_s\|^2}{2}\cos{2\alpha} & {\frac{\|\boldsymbol\psi_s\|^2}{2}\sin{2\alpha}} \\
    {\frac{\|\boldsymbol\psi_s\|^2}{2}\sin{2\alpha}} & Q - \frac{\|\boldsymbol\psi_s\|^2}{2}\cos{2\alpha}
    \end{pmatrix},
    \label{eq:ellipse-covariance}
\end{aligned}
\end{equation}
where $\alpha = \tan^{-1}(\psi_y(s)/\psi_x(s))$, $Q = \sqrt{4A^2 + \|\boldsymbol\psi_s\|^4\pi^2}/2\pi$, and $\lambda$ is a scaling parameter that controls the overall intensity of the covariance. 
Fig.~\ref{fig:focus-points-exp} shows a series of randomly generated focus points and their resulting ellipses. 
This demonstrates how the spatially distributed pairs $\boldsymbol{\psi}_s \coloneqq (\psi_x(s), \psi_y(s))$ give rise to a spatially distributed covariance matrix $\Sigma_s$.
The derivation of \eqref{eq:ellipse-covariance} can be found in Appendix~\ref{append:reparametrization-gaussian}.

\begin{figure}[!t]
\centering
\includegraphics[width=\linewidth]{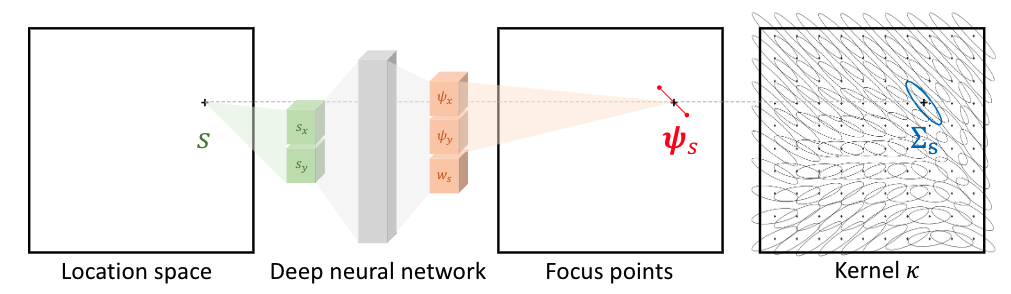}
\caption{An illustration of the deep neural network that maps an arbitrary spatial location $s$ to its covariance $\Sigma_s$ and the corresponding weight $w_s$.}
\label{fig:kernel-nn-illustration}
\end{figure}

\begin{figure}[!t]
\centering
\includegraphics[width=\linewidth]{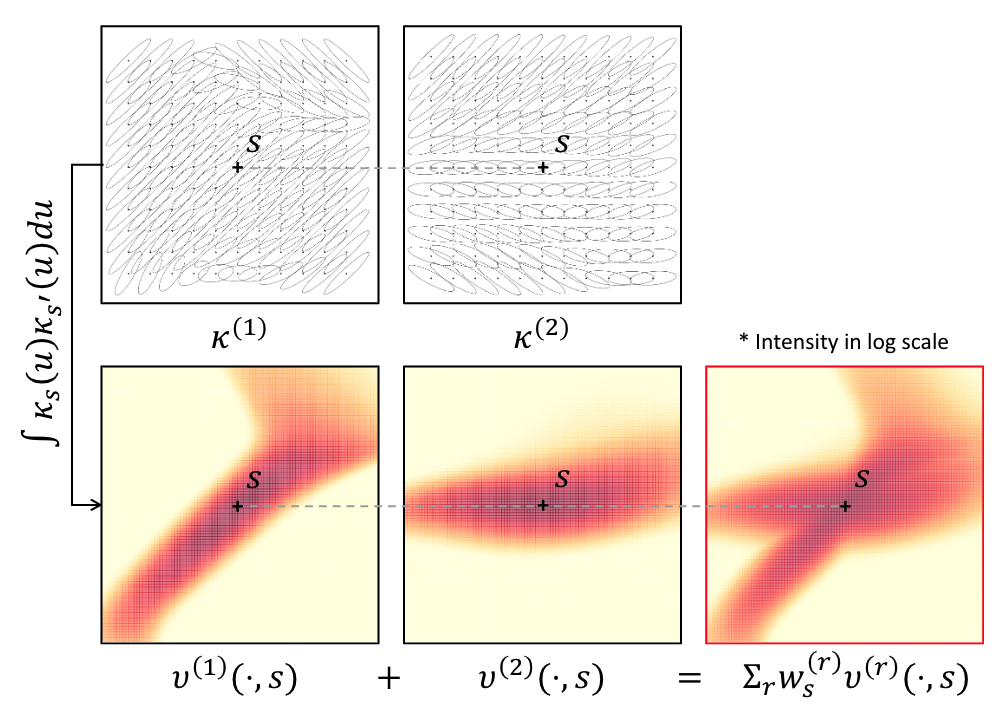}
\caption{An examples of the spatial kernel with two components $\sum_r w_{s}^{(r)} \upsilon^{(r)}(\cdot, s)$ evaluated at the same location $s$. This instance is constructed using two different kernel $\kappa$, which are parameterized by two randomly generated $\varphi_1$ and $\varphi_2$.}
\label{fig:kernel-instances}
\end{figure}

\vspace{.1in}
\noindent\emph{Neural network representation for focus points.}
Here we represent the mapping $\varphi: \mathcal{S} \to \Psi \times [0, 1]$ from the location space $\mathcal{S}$ to the joint space of focus point $\Psi$ and the weight $[0, 1]$ using a deep neural network. To be specific, the input of the network is the location $s$, and the output of the networks is the concatenation of the corresponding focus points $\boldsymbol{\psi}_s$ of that location and the weight $w_s$ defined in \eqref{eq:spatio-temporal-kernel}. The architecture of the neural network has been described in Fig.~\ref{fig:kernel-nn-illustration}. 
In Fig.~\ref{fig:kernel-instances}, we also demonstrate two specific instances of the resulting spatial kernel $\upsilon$ given two different $\kappa$. 
This implies that the neural network $\varphi$ encodes the non-homogeneous geographical information across the region in spreading the virus.

\section{Efficient Computation for Large-scale Data Set}
\label{sec:efficient-computation}

There are two major challenges in learning the model and calculating the objective defined in \eqref{eq:objective}.
First, the GP approach is notoriously intractable for large data sets since the computations require the inversion of a matrix of size $N \times N$, which scales as $O(N^3)$ \cite{rasmussen2003gaussian}. 
In this study, the data set includes 3,144 counties and more than 50 weeks extending from March 2020 to January 2021 ($N = 3,144 \times 50$).
Second, the inference of the posterior distribution of the hotspot $p(\mathbf f|\mathbf{h})$ requires the calculation of integral $\int p(\mathbf{h}|\mathbf{f}) p(\mathbf{f}) d\mathbf{f}$, which is an intractable integration. 

To circumvent these two issues, we derive sparse models for both cases and hotspots similar to \cite{titsias2009variational, hensman2013gaussian, hensman2015scalable}, where their log marginal likelihood is computationally tractable for large data sets and they do not require an analytical expression for inferring the non-Gaussian posterior distribution. First, we define a small set of inducing variables that aim to best approximate the training data. Then we adopt a variational learning strategy for such sparse approximation and jointly infer the inducing inputs and other model parameters by maximizing a lower bound of the true log marginal likelihood \cite{titsias2009variational, hensman2015scalable}.
Since the learning strategy can be applied to the above two models, we use $\mathbf{y}$ to represent both cases and hotspots for notational brevity in the following discussion.
Lastly, the objective is jointly learned by performing stochastic gradient descent.

\subsection{Variational Inference for Sparse Gaussian Process}
\label{sec:variational-inference}

\begin{figure}[!t]
\centering
\includegraphics[width=.8\linewidth]{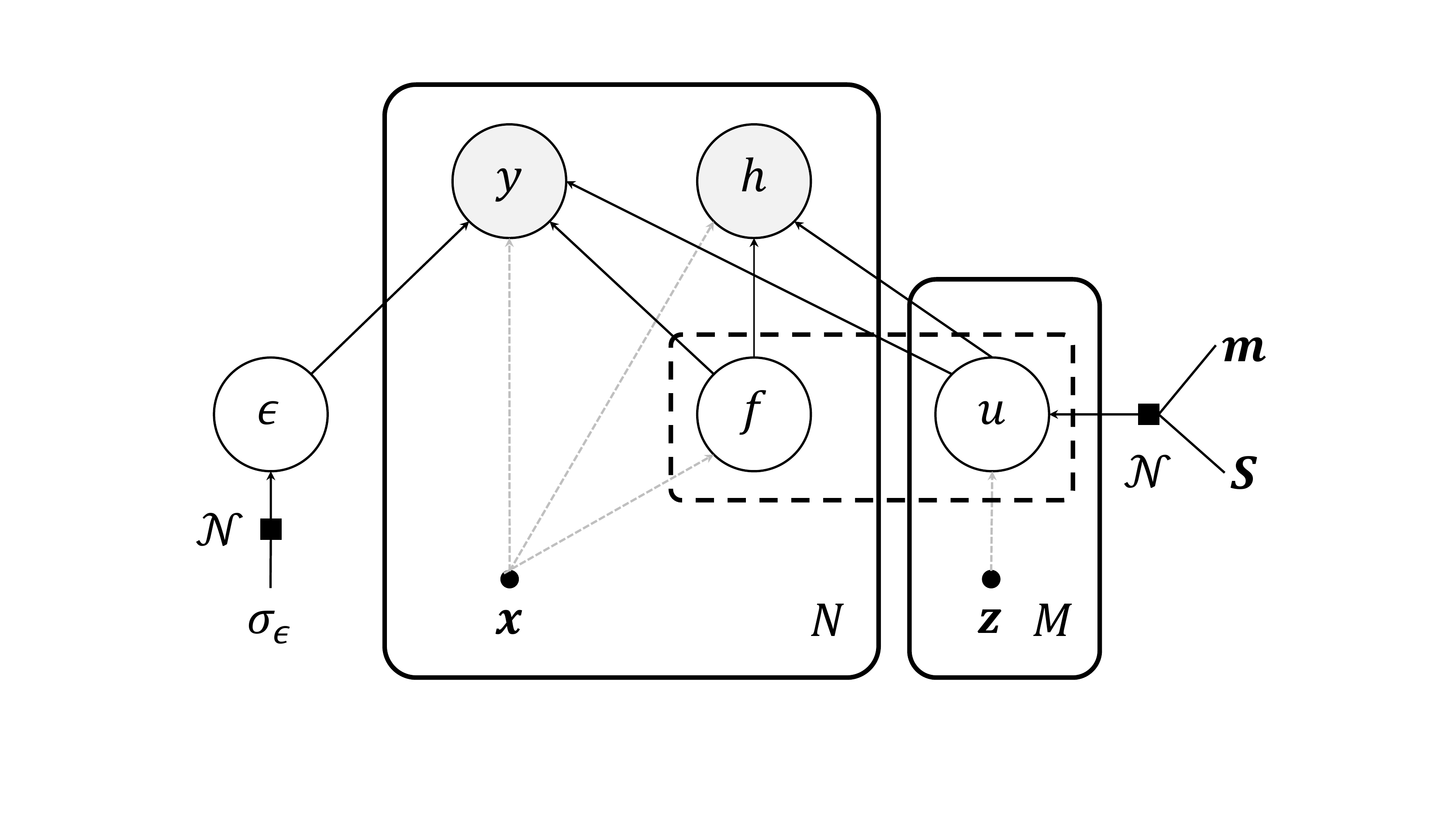}
\caption{A diagram of our graphical model. The gray and white nodes represent the observed and latent variables, respectively; the black dots represent the input variables; the black boxes represent the prior distribution. We use the dashed box to highlight the joint distribution of $\mathbf{f}$ and $\mathbf{u}$ defined in \eqref{eq:joint-dist-f-u}.}
\label{fig:model-illustration}
\end{figure}

Unlike the exact GP approaches approximating the true covariance by the Nystr\"{o}m approximation \cite{rasmussen2003gaussian},
we desire a sparse method that directly approximates the posterior GP's mean and covariance function.
Now we introduce a small set of $M$ auxiliary inducing variables $\mathbf{u}$ evaluated at the pseudo-inputs $\mathbf{Z} \coloneqq \{\mathbf{z} \in \mathscr{X}\}$;
$\mathbf{Z}$ can be a subset of the training inputs or auxiliary pseudo-points \cite{snelson2005sparse}.
$\mathbf{u}$ are function points drawn from the same GP prior as the training functions $\mathbf{f}$ in \eqref{eq:prior}, so the joint distribution can be written as
\begin{equation}
    p([\mathbf{f}, \mathbf{u}]^\top) = \mathcal{N}\left(
    \mathbf{0},~
    \begin{bmatrix}
    \mathbf{K}_{XX} & \mathbf{K}_{XZ} \\
    \mathbf{K}_{XZ}^\top & \mathbf{K}_{ZZ}
    \end{bmatrix}
    \right),
    \label{eq:joint-dist-f-u}
\end{equation}
where $\mathbf{K}_{ZZ}$ is formed by evaluating the kernel function pairwisely at all pairs of inducing points in $\mathbf{Z}$, and $\mathbf{K}_{XZ}$ is formed by evaluating the kernel function across the data points $\mathbf{X}$ and inducing points $\mathbf{Z}$ similarly. 
Fig.~\ref{fig:model-illustration} presents the diagram of our graphical model, consisting of observed variables $\mathbf{y}, \mathbf{h}$, latent variable $\mathbf{f}$, and the introduced auxiliary variable $\mathbf{z}$.

To obtain computationally efficient inference, we approximate the posterior distribution $p(\mathbf{f}, \mathbf{u}|\mathbf{y})$ over random variable vector $\mathbf{f}$ and $\mathbf{u}$ by a variational distribution $q(\mathbf{f}, \mathbf{u})$. 
We assume this variational distribution $q(\mathbf{f}, \mathbf{u})$ can be factorized as
$
    q(\mathbf{f}, \mathbf{u}) 
    = p(\mathbf{f} | \mathbf{u}) q(\mathbf{u})
$.
To jointly determine the variational parameters and model parameters, the variational evidence lower bound (ELBO) substitutes for the marginal likelihood $\ell_y(\boldsymbol{\theta})$ and $\ell_h(\boldsymbol{\theta})$ defined in \eqref{eq:objective}:
\begin{equation}
    \log p(\mathbf{y}) \ge \mathbb{E}_{q(\mathbf{f})}\left [ \log p(\mathbf{y}|\mathbf{f}) \right ] - \text{KL}\left [ q(\mathbf{u}) || p(\mathbf{u}) \right ],
    \label{eq:elbo}
\end{equation}
where $\text{KL}[q||p]$ denotes the Kullback–Leibler (KL) divergence between two distributions $q$ and $p$ \cite{kullback1951information}. We have defined: $q(\mathbf{f}) \coloneqq \int p(\mathbf{f}|\mathbf{u}) q(\mathbf{u}) d\mathbf{u}$ and assume $q(\mathbf{u}) \coloneqq \mathcal{N}(\mathbf{m}, \mathbf{S})$, which is the most common way to parameterize the prior distribution of inducing variables in terms of a mean vector $\mathbf{m}$ and a covariance matrix $\mathbf{S}$. To ensure that the covariance matrix remains positive definite, we represent it
using a lower triangular form $\mathbf{S} = \mathbf{L} \mathbf{L}^\top$. 
This leads to the following analytical form for $q(\mathbf{f})$:
\[
    q(\mathbf{f}) = \mathcal{N}(\mathbf{A}\mathbf{m}, \mathbf{K}_{XX} + \mathbf{A} (\mathbf{S} - \mathbf{K}_{ZZ}) \mathbf{A}^\top),
\]
where $\mathbf{A} = \mathbf{K}_{XZ} \mathbf{K}_{ZZ}^{-1}$. 
In classification or regression, we also factorize the likelihood as $p(\mathbf{y}|\mathbf{f}) = \prod_{n=1}^N p(y_n|f_n)$ for the ease of computation in \eqref{eq:elbo}. 
Therefore, the ELBO objective can be rewritten as
\begin{equation}
\begin{aligned}
    &~\ell_\text{ELBO}(\boldsymbol{\theta}, \mathbf{Z}, \mathbf{m}, \mathbf{S}) \coloneqq \\
    &~\sum_{n=1}^N \mathbb{E}_{q(f_n)}\left [ \log p(y_n|f_n) \right ] - \text{KL}\left [ q(\mathbf{u}) || p(\mathbf{u}) \right ].
    \label{eq:obj-elbo}
\end{aligned}
\end{equation}
In practice, the one dimensional integrals of the log-likelihood in \eqref{eq:obj-elbo} can
be computed by Gauss-Hermite quadrature \cite{liu1994note}.
In contrast to directly maximizing the marginal log likelihood defined in \eqref{eq:objective}, computing this objective and its derivatives can be done in $O(NM^2)$ time. The derivation of the ELBO can be found in Appendix~\ref{append:elbo}.

\subsection{Prediction with Variational Posterior}

To make one-week ahead predictions for the hotspots and the number of confirmed cases, we first need to derive the posterior distribution of prediction $p(\mathbf{f}|\mathbf{y}, \mathbf{h})$ given the past observation.
Suppose we have the spatio-temporal coordinates $\mathbf{X}_t \coloneqq \{\mathbf{x}_{j\tau}\}_{j\in\mathscr{I}, \tau\le t}$ and their observations $\mathbf{y}_t \coloneqq \{y_{j\tau}\}_{j\in\mathscr{I}, \tau\le t}$, $\mathbf{h}_t \coloneqq \{h_{j\tau}\}_{j\in\mathscr{I}, \tau\le t}$ until time $t$ and the optimal inducing points $\mathbf{Z}$.
We assume that the unobserved future data comes from the same generation process.
Therefore, for all the locations at time $t+1$, i.e., $\mathbf X_* \coloneqq \{\mathbf{x}_{j,t+1}\}_{j\in\mathscr{I}}$, we first estimate their means according to \eqref{eq:mean} denoted by $\boldsymbol{\mu}_* \coloneqq \{\mu_{j,t+1}\}_{j\in\mathscr{I}}$, then the distribution of one-week-ahead prediction $\mathbf f_* \coloneqq \{\hat f_{j, t+1}\}_{j\in\mathscr{I}}$ is given by 
\begin{equation}
    p(\mathbf f_* | \mathbf{y}_t, \mathbf{h}_t) 
    = \mathcal{N}(\mathbf{A}_* \mathbf{m}, \mathbf{A}_* \mathbf{S} \mathbf{A}_*^\top + \mathbf{B}_*),
    \label{eq:pred-posterior}
\end{equation}
where $\mathbf{A}_* = \mathbf{K}_{*Z}\mathbf{K}_{ZZ}^{-1}$ and $\mathbf{B}_* = \mathbf{K}_{**} - \mathbf{K}_{*Z} \mathbf{K}_{ZZ}^{-1} \mathbf{K}_{*Z}^\top$.
The $\mathbf{K}_{* Z}$ denotes a $I \times M$ matrix and its entries are pairwise evaluations of $k(\mathbf{x}_*, \mathbf{z})$ where $\mathbf{x}_* \in \mathbf{X}_*$ and $\mathbf{z} \in \mathbf{Z}$. The derivation of the predictive posterior can be found in Appendix~\ref{append:pred-posterior}. The prediction for the number of cases and the probability of hotspots therefore can be made by plugging \eqref{eq:pred-posterior} into \eqref{eq:y-conditional-f} and \eqref{eq:h-conditional-f}, respectively.
We consider that a detector would raise an alarm if the hotspot probability $h_{it}$ for location $i$ at time $t$ is above the pre-set threshold $\zeta_i$. This threshold is chosen for each location by a grid search in $[0, 1]$. For each location, the threshold with the largest in-sample $F_1$ score is chosen.

The above formula reveals that predictive posterior distribution only depends on the inducing variables $\mathbf{u}$ at learned spatio-temporal coordinates $\mathbf{Z}$ and does not depend on the $\mathbf{f}$ at training coordinates. 
This shows that all the information from the training data has been summarized by the proposed posterior distribution $q(\mathbf{u})$ defined in Section~\ref{sec:variational-inference} and the prediction for future weeks can be carried out efficiently.

\subsection{Stochastic Gradient Descent based Optimization}

Now we describe our learning algorithm. The optimal parameters of the proposed model can be found by maximizing the combined objective \eqref{eq:objective} using gradient-based optimization.
However, the full gradient evaluation can still be expensive to carry out. 
With a sparse prior (inducing variables), even though we can tackle the computational challenge in inverting a big matrix, evaluating the gradient of the first term in \eqref{eq:obj-elbo} still requires the full data set, which is memory-intensive if the size of the data set $N$ is too large.
To alleviate the problem of expensive gradient evaluation, we adopt a stochastic gradient-based method \cite{hensman2013gaussian} and only compute the gradient of the objective function with respect to the GP's parameters denoted by $\boldsymbol{\theta} \coloneqq \{\{\varphi^{(r)}\}_{r=1,\dots,R}, \sigma_\nu\}$, evaluated on a random subset of the data at each iteration. 

Additionally, the conventional stochastic gradient descent algorithm assumes that the parameters' loss geometry is Euclidean. 
This is a non-ideal assumption for the parameters of many distributions, e.g., Gaussian. 
Here we follow the idea of \cite{hensman2013gaussian, martens2014new} and apply adapted stochastic gradient descent to the variational parameters ($\mathbf{Z}, \mathbf{m}, \mathbf{S}$) in our GP model by taking steps in the direction of the approximate natural gradient.
These gradients are computed by the usual gradient re-scaled by the inverse Fisher information. The Kullback–Leibler divergence is used to measure the ``closeness'' in the variational distribution space.
Our learning algorithm is summarized in Algorithm~\ref{algo:learning}.

\begin{algorithm}[!t]
\SetAlgoLined
    {\bfseries Initialization:} Randomly initialize $\boldsymbol{\theta}, \mathbf{Z}, \mathbf{m}, \mathbf{S}$\;
    {\bfseries Input:} Data set $\mathbf{X}, \mathbf{y}, \mathbf{h}$; Number of iterations $B$; Batch size $n$\;
    \For{$b = \{1, \dots, B \}$}{
        Sample a subset $\mathbf{X}_b, \mathbf{y}_b, \mathbf{h}_b$ with $n$ points from $\mathbf{X}, \mathbf{y}, \mathbf{h}$, respectively\; 
        Calculate ELBO of $\ell_h$ and $\ell_y$ based on \eqref{eq:obj-elbo} given data $\mathbf{X}_b, \mathbf{y}_b, \mathbf{h}_b$\;
        Calculate the gradient of \eqref{eq:objective} {\it w.r.t.} $\boldsymbol{\theta}$\;
        Calculate the natural gradient of \eqref{eq:objective} {\it w.r.t.} $\mathbf{Z}, \mathbf{m}, \mathbf{S}$\;
        Ascend the gradient of $\boldsymbol{\theta}, \mathbf{Z}, \mathbf{m}, \mathbf{S}$\;
    }
\caption{Learning algorithm for the COVID-19 hotspot detection framework}
\label{algo:learning}
\end{algorithm}

\begin{figure*}[!t]
\centering
\begin{subfigure}[h]{.49\linewidth}
\includegraphics[width=\linewidth, frame]{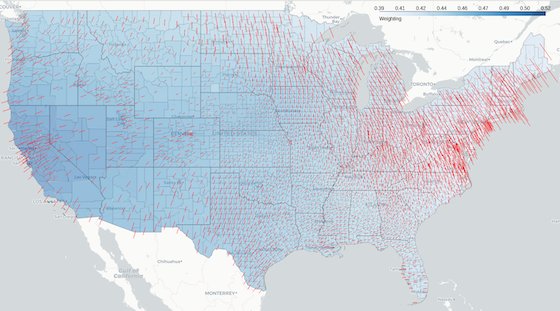}
\caption{$\kappa^{(1)}_s$}
\end{subfigure}
\begin{subfigure}[h]{.49\linewidth}
\includegraphics[width=\linewidth, frame]{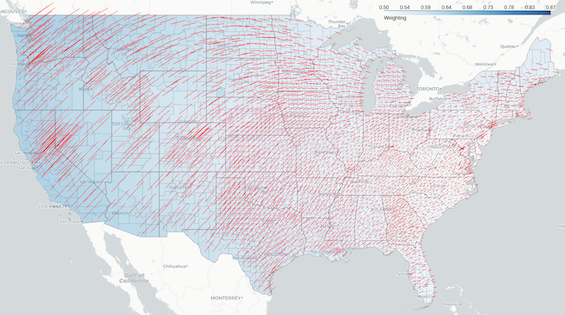}
\caption{$\kappa^{(2)}_s$}
\end{subfigure}
\vfill
\begin{subfigure}[h]{.49\linewidth}
\includegraphics[width=\linewidth, frame]{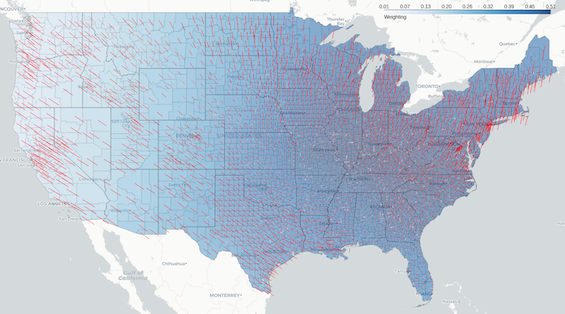}
\caption{$\kappa^{(3)}_s$}
\end{subfigure}
\begin{subfigure}[h]{.49\linewidth}
\includegraphics[width=\linewidth, frame]{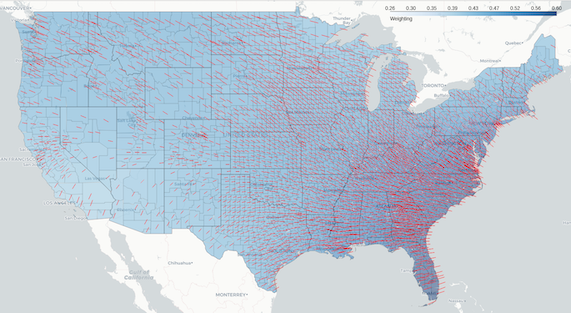}
\caption{$\kappa^{(4)}_s$}
\end{subfigure}
\caption{Visualizations of the learned kernel induced feature $\kappa^{(r)}_s$ using COVID-19 data set. Each panel shows one of four kernel components, where the line segment is the edge that connects two focus points of $s$, indicating the shape and the rotation of the kernel at that location; the shaded area shows the intensity of the corresponding weight $w_{s}^{(r)}$ at location $s$; the darker the region, the larger the weight.}
\label{fig:kernel-component}
\end{figure*}

\begin{figure*}[!t]
\centering
\begin{subfigure}[h]{0.245\linewidth}
\includegraphics[width=\linewidth, frame]{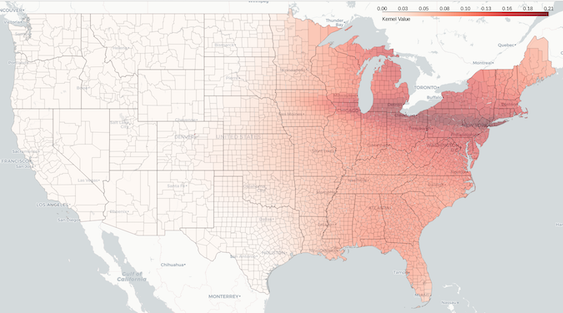}
\caption{New York}
\end{subfigure}
\begin{subfigure}[h]{0.245\linewidth}
\includegraphics[width=\linewidth, frame]{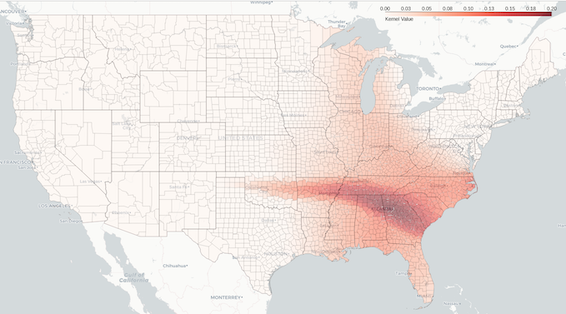}
\caption{Atlanta}
\end{subfigure}
\begin{subfigure}[h]{0.245\linewidth}
\includegraphics[width=\linewidth, frame]{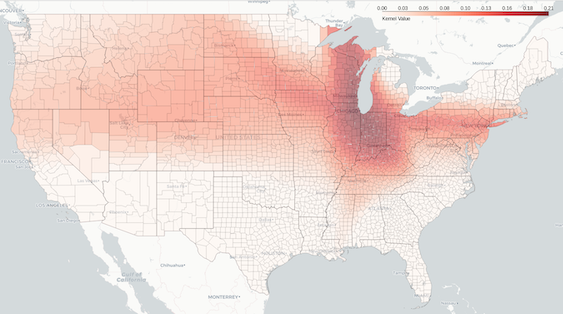}
\caption{Chicago}
\end{subfigure}
\begin{subfigure}[h]{0.245\linewidth}
\includegraphics[width=\linewidth, frame]{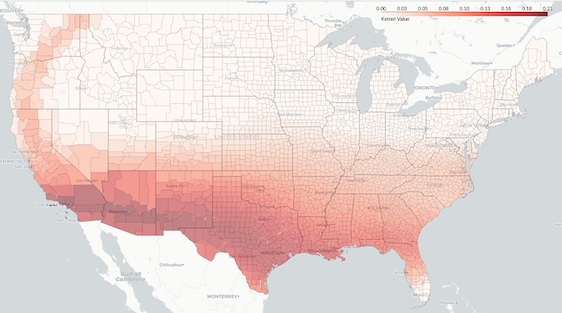}
\caption{Los Angeles}
\end{subfigure}
\caption{Examples of the learned spatial kernel $\sum_{r=1}^R w_{s}^{(r)} \upsilon^{(r)}(\cdot, s)$ with four components evaluated at four major metropolitan areas in the U.S.. These maps show the spatial influence of these area to other region of the U.S.. The color depth indicates the intensity of the kernel value; the darker the color the higher the kernel's value.}
\label{fig:kernel}
\end{figure*}

\section{Results}
\label{sec:results}

This section reports the numerical results of our study. 
In the following examples, we consider the spatial kernel with $R=4$ components and fit the model with $M=500$ inducing variables using the COVID-19 data set described in Section~\ref{sec:data}. 
For each spatial kernel component $r$, 
we choose a three-layer neural network with 64 nodes per layer to represent its mapping $\varphi^{(r)}$ through cross-validation (Appendix~\ref{append:cv-nn-res}).
We first evaluate the explanatory power of the proposed framework by investigating the learned spatial kernel function using COVID-19 data. 
We demonstrate the interpretable components of our model and visualize the spatio-temporal correlation across regions discovered by our fitted model.
Then we examine the result of the hotspot detection and the case prediction by visualizing the one-week-ahead predictions and their distribution.
We emphasize that our model not only generates accurate predictions, but also quantifies the uncertainty about the predictions.
Lastly, we compare our method with six other commonly-used binary classification approaches by evaluating their out-of-sample predictive performance. 
The inputs to the hotspot prediction model are past county-level case and death records, identified hotspots, and community mobility information. 
For ease of presentation, we only focus on the counties in the contiguous United States. 

\subsection{Model Interpretation}

\begin{figure*}[!h]
\centering
\begin{subfigure}[h]{0.24\linewidth}
\includegraphics[width=\linewidth]{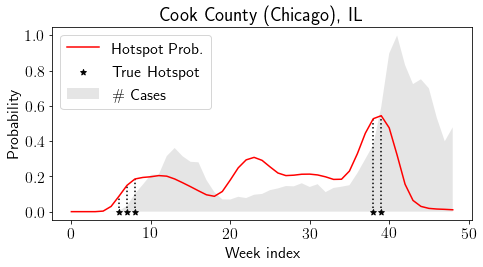}
\end{subfigure}
\begin{subfigure}[h]{0.24\linewidth}
\includegraphics[width=\linewidth]{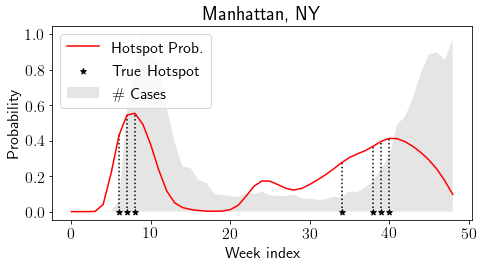}
\end{subfigure}
\begin{subfigure}[h]{0.24\linewidth}
\includegraphics[width=\linewidth]{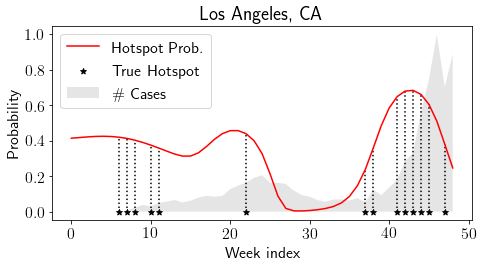}
\end{subfigure}
\begin{subfigure}[h]{0.24\linewidth}
\includegraphics[width=\linewidth]{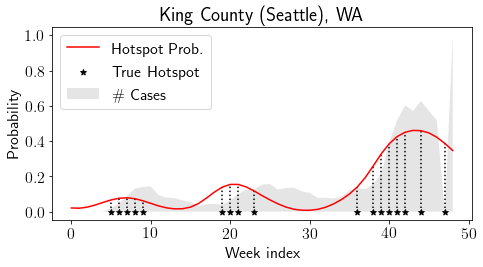}
\end{subfigure}
\vfill
\begin{subfigure}[h]{0.24\linewidth}
\includegraphics[width=\linewidth]{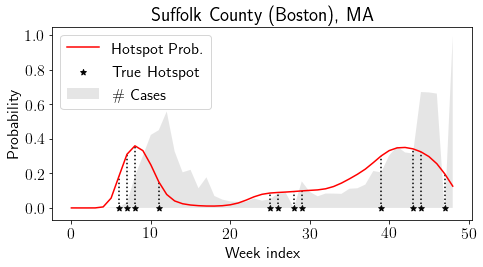}
\end{subfigure}
\begin{subfigure}[h]{0.24\linewidth}
\includegraphics[width=\linewidth]{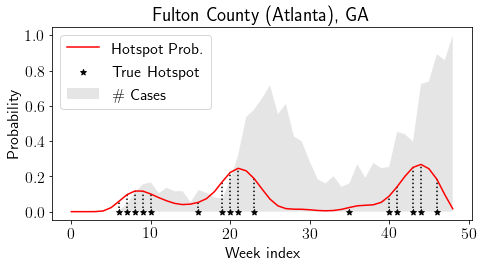}
\end{subfigure}
\begin{subfigure}[h]{0.24\linewidth}
\includegraphics[width=\linewidth]{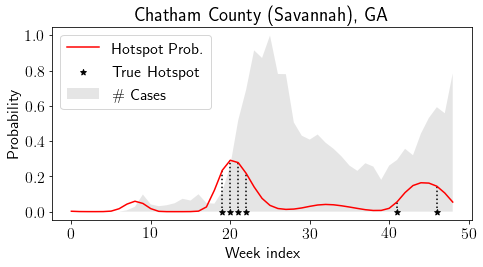}
\end{subfigure}
\begin{subfigure}[h]{0.24\linewidth}
\includegraphics[width=\linewidth]{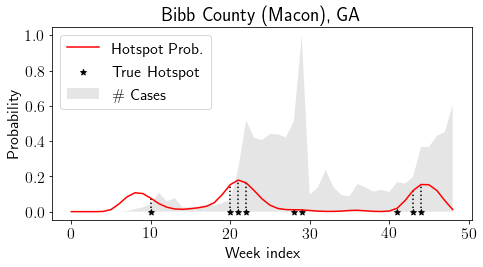}
\end{subfigure}
\caption{Temporal view of one-week-ahead and county-wise hotspot probability $p(\mathbf{h}_*)$ suggested by our fitted model ($\delta = 10^{-5}$) using COVID-19 data. The first 6 panels represent major metropolitan areas and the last two panels represent less populated counties in the United States.}
\label{fig:hotspot-time}
\end{figure*}

\begin{figure*}[!h]
\centering
\begin{subfigure}[h]{0.24\linewidth}
\includegraphics[width=\linewidth, frame]{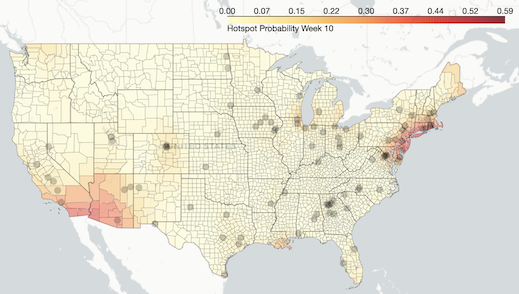}
\caption{April 12, 2020}
\end{subfigure}
\begin{subfigure}[h]{0.24\linewidth}
\includegraphics[width=\linewidth, frame]{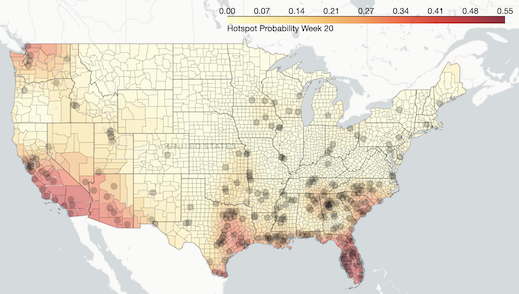}
\caption{August 9, 2020}
\end{subfigure}
\begin{subfigure}[h]{0.24\linewidth}
\includegraphics[width=\linewidth, frame]{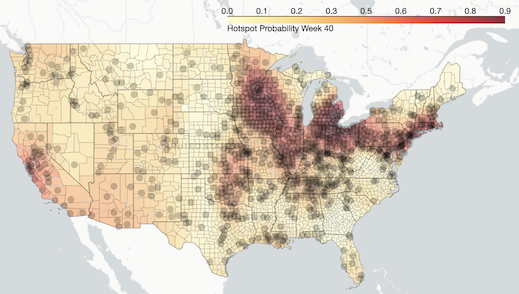}
\caption{October 4, 2020}
\end{subfigure}
\begin{subfigure}[h]{0.24\linewidth}
\includegraphics[width=\linewidth, frame]{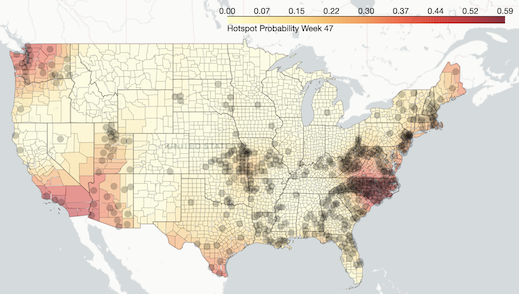}
\caption{December 6, 2020}
\end{subfigure}
\caption{Spatial view of one-week-ahead and county-wise hotspot probability $p(\mathbf{h}_*)$ suggested by our fitted model ($\delta = 10^{-5}$) using real COVID-19 data. This figure presents examples at four particular weeks, where the color depth indicates the probability of predicted hotspots and the black circles represent the hotspots given in the data.}
\label{fig:hotspot-map}
\end{figure*}

The proposed framework offers a unique opportunity for understanding the dynamics of the spread of COVID-19 utilizing the carefully crafted kernel design. In this experiment, we fit the model using the entire COVID-19 data and then visually examine the learned spatial kernel. 

We first visualize the learned kernel induced feature $\kappa_s^{(r)}$ of each spatial kernel component in Fig.~\ref{fig:kernel-component}, which portrays the spatial pattern of virus' propagation.
Recall that, for any arbitrary $s$, $\kappa^{(r)}_s$ is a normal kernel centered at $s$ with spatially varying covariance matrix $\Sigma^{(r)}_s$, which can be uniquely represented by its focus points. 
Here, we connect two focus points with a red line at each location and plot them on the map. 
Length and rotation angle of the red line at $s$ represents the strength and direction of the influence of location $s$, respectively, jointly determining the shape of its covariance matrix $\Sigma^{(r)}_s$. The color depth indicates the weight $w_s^{(r)}$, representing the ``significance'' of location $s$ in the kernel component $r$. 

To intuitively interpret the learned spatial kernel, we also visualize the kernel evaluation given one of its inputs, i.e., $\sum_{r=1}^R w_{s}^{(r)} \upsilon^{(r)}(s, \cdot)$.
Such kernel evaluation represents the spatial correlation (or sphere of influence) of a particular location spreading the virus.
The area in darker red signals that the corresponding counties are more likely to become the next hotspot if an outbreak is detected in the region of interest. 
In Fig.~\ref{fig:kernel}, we observe that these major metropolitan areas have a substantially different spatial correlation with their neighboring regions due to the non-stationarity of the spatial kernel. 
For example,
as one of the nation’s major economic and transportation hubs, 
New York has a significant impact on the entire Eastern United States, while Atlanta only has a regional influence in the Southeastern United States. 
Chicago and Los Angeles, the second and third most populous cities in the United States, can extend their influences to the entire north and south of the country, respectively.
Such interpretability of our spatial kernel function could be of particular importance for the policymakers or the local authorities, suggesting that counties with more extensive influence on other regions should receive more attention in epidemic prevention. 
We also note that increasing the number of spatial kernel components could further enhance the flexibility and the interpretability of the model (Appendix~\ref{append:comp-r-exp-results}); however, due to the need for additional parameters in the neural networks, the computational time dramatically increases when $R \ge 3$, with minimal performance improvement.

\subsection{Hotspot Detection}

We evaluate the one-week-ahead in-sample prediction accuracy of our proposed hotspot detection framework at the county level. We first fit the model using the entire data set from March 15, 2020, to January 17, 2021, which contains 3,144 counties and 50 weeks in total. The in-sample prediction for time $t$ is obtained by feeding the data before $t$ into the fitted model and predicting the one-week-ahead hotspots. 
We test our model with different $\delta$ values and perform cross-validation to identify the optimal $\delta$.
In Fig.~\ref{fig:hotspot-time}, we report the in-sample prediction results for eight representative locations, which include six major metropolitan cities and two sparsely populated counties. 
The shaded gray area indicates the number of cases reported in that location, and the black star indicates the time of the identified hotspot. 
The solid red line represents the corresponding estimated hotspot probability. 
The hotspot probability resulting from our model is considerably high whenever a genuine hotspot occurs and considerably low otherwise, which confirms the effectiveness of our framework. 
In Fig.~\ref{fig:hotspot-map}, we visualize the prediction results on the map to intuitively examine the predictive performance from the spatial perspective. 
We select four particular weeks representing different stages of the COVID-19 pandemic in 2020. The black dot indicates the genuinely identified hotspot, and the color depth indicates the hotspot probability suggested by our fitted model.
As we can see, our method can capture the spatial occurrence of these hotspots nicely, in which regions with sparsely distributed hotspots usually have a lower probability. In comparison, other regions with densely distributed hotspots have a higher probability. 
We emphasize that our hotspot detection framework can provide a realistic prediction that varies smoothly over time and space due to our GP assumption. 
This can be extremely useful when we try to make a continuous prediction or estimate the likelihood of a hotspot to happen at an arbitrary spatio-temporal coordinate. 

\subsection{Case Prediction and Uncertainty Quantification}

\begin{figure*}[!h]
\centering
\begin{subfigure}[h]{0.24\linewidth}
\includegraphics[width=\linewidth]{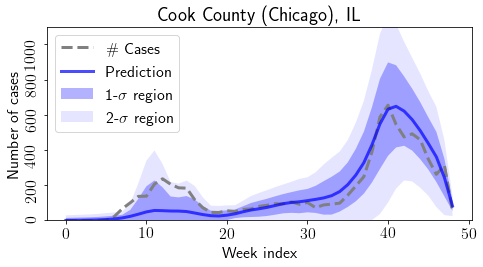}
\end{subfigure}
\begin{subfigure}[h]{0.24\linewidth}
\includegraphics[width=\linewidth]{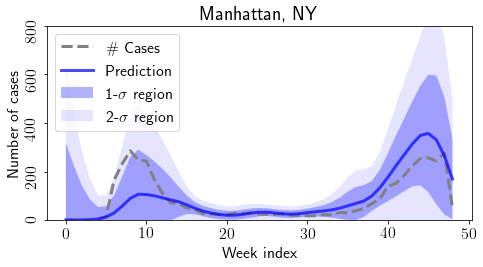}
\end{subfigure}
\begin{subfigure}[h]{0.24\linewidth}
\includegraphics[width=\linewidth]{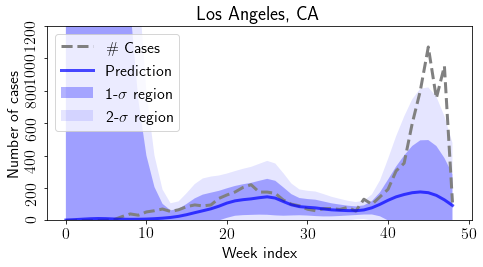}
\end{subfigure}
\begin{subfigure}[h]{0.24\linewidth}
\includegraphics[width=\linewidth]{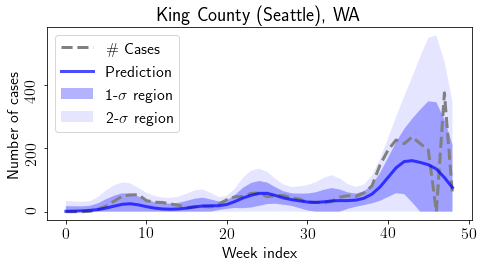}
\end{subfigure}
\vfill
\begin{subfigure}[h]{0.24\linewidth}
\includegraphics[width=\linewidth]{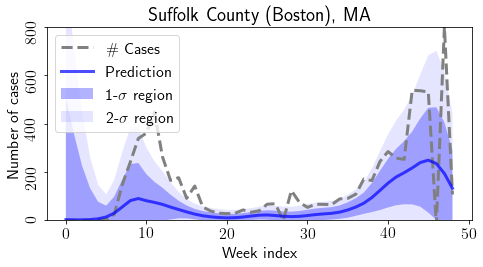}
\end{subfigure}
\begin{subfigure}[h]{0.24\linewidth}
\includegraphics[width=\linewidth]{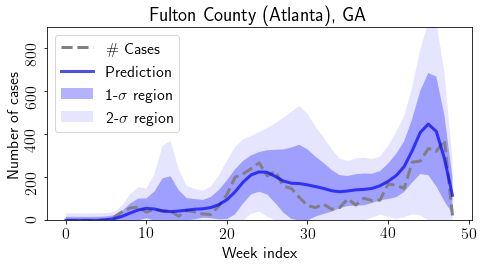}
\end{subfigure}
\begin{subfigure}[h]{0.24\linewidth}
\includegraphics[width=\linewidth]{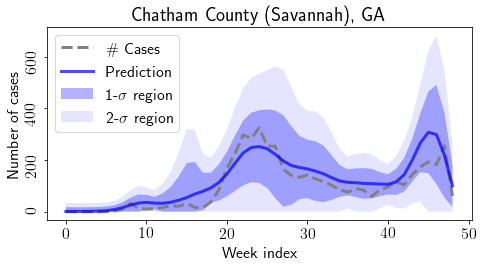}
\end{subfigure}
\begin{subfigure}[h]{0.24\linewidth}
\includegraphics[width=\linewidth]{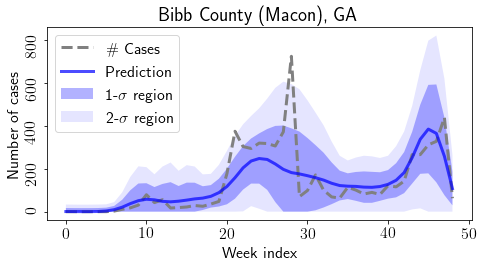}
\end{subfigure}
\caption{Temporal view of one-week-ahead and county-wise case prediction $\mathbf{y}_*$ suggested by our fitted model ($\delta = 10^{-5}$). This figure presents eight examples for major metropolitan area (first six panels) and less populated counties (last two panels) in the United States.}
\label{fig:case-prediction}
\end{figure*}

\begin{figure*}[!h]
\centering
\begin{subfigure}[h]{0.245\linewidth}
\includegraphics[width=\linewidth, frame]{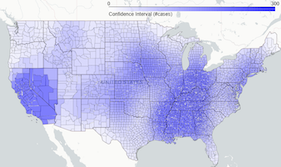}
\caption{April 12, 2020}
\end{subfigure}
\begin{subfigure}[h]{0.245\linewidth}
\includegraphics[width=\linewidth, frame]{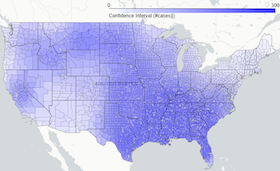}
\caption{August 9, 2020}
\end{subfigure}
\begin{subfigure}[h]{0.245\linewidth}
\includegraphics[width=\linewidth, frame]{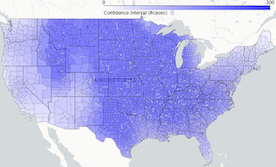}
\caption{October 4, 2020}
\end{subfigure}
\begin{subfigure}[h]{0.245\linewidth}
\includegraphics[width=\linewidth, frame]{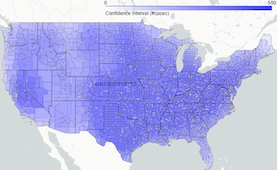}
\caption{December 6, 2020}
\end{subfigure}
\caption{Spatial view of the confidence interval of one-week-ahead and county-wise case prediction $\mathbf{y}_*$ at four particular weeks. The color depth indicates the length of 95\% confidence interval of the prediction; the darker the region, the more uncertain the prediction becomes.}
\label{fig:uncertainty-map}
\end{figure*}

Our proposed framework also provides case prediction and uncertainty quantification besides hotspot detection.
In Fig.~\ref{fig:case-prediction}, we present the predicted case number $\mathbf{y}_*$ over time as well as its confidence interval for the eight same locations that appeared previously. 
The black dash line represents the real reported cases, and the solid blue line represents the prediction $\mathbf{y}_*$ suggested by our case model. The one and two-$\sigma$ regions are highlighted by the dark and light blue shaded areas. 
As we can see, the prediction result captures the general trend of the case records, which confirms that the case model can successfully extract useful information from the cases that will be used to regularize the hotspot model by optimizing \eqref{eq:objective}.
We note that the estimated confidence interval reflects the uncertainty level of our prediction for both cases and hotspots since the confidence interval only depends on the latent spatio-temporal variable $\mathbf{f}$. 
In Fig.~\ref{fig:uncertainty-map}, we show the confidence interval over the map for four different weeks. The color depth indicates the uncertainty level (the length of the confidence interval) for that location. This intuitively tells us which area we are confident in making predictions and how this confidence changes over space. 

\subsection{Comparison with Baselines}\label{sec:compare_baseline_detect}

We adopt standard performance metrics, including precision, recall, and $F_1$ score. This choice is because hotspot detection can be viewed as a binary classification problem. We aim to identify a hotspot for a particular location at a particular week in the data. 
Define the set of all identified hotspots as $U$, the set of detected hotspots using our method as $V$.
Then precision $P$ and recall $R$ are defined as
$
    P = |U \cap V|/|V|,~R = |U \cap V|/|U|,
$
where $|\cdot|$ is the number of elements in the set. 
The $F_1$ score combines the \emph{precision} and \emph{recall}: $F_1 = 2 P R / (P + R)$ and the higher $F_1$ score the better. Since numbers of hotspots in real data are highly sparse (comparing to the total number of spatio-temporal coordinates), we do not use the ROC curve (true positive rate versus false-positive rate) in our setting. 
The evaluation procedure is described as follows. Given the observed hotspot and other covariates (cases, deaths, and mobility) until $t$, we perform detection for all the locations at time $t+1$. If the detected hotspot were indeed identified as a genuine hotspot by CDC, then it is a success. Otherwise, it is a misdetection. In our data, there are $50 \times 3144 = 157,200$ spatio-temporal coordinates in total, and 12,000 of them were identified as genuine hotspots.

\begin{table}[!h]
\centering
\caption{$F_1$ score of out-of-sample hotspot detections.}
\label{tab:f1score}
\resizebox{.8\linewidth}{!}{%
    \begin{threeparttable}
        \centering
        \begin{tabular}{l:lll}
            \toprule[1pt]\midrule[0.3pt]
            \multicolumn{1}{c:}{} & \multicolumn{1}{c}{Precision} & \multicolumn{1}{c}{Recall} & \multicolumn{1}{c}{$F_1$ score}\\ \hline
            \texttt{Perceptron} & \multicolumn{1}{c}{0.424} & \multicolumn{1}{c}{0.242} & \multicolumn{1}{c}{0.308}\\
            \texttt{Logistic} & \multicolumn{1}{c}{0.564} & \multicolumn{1}{c}{0.178} & \multicolumn{1}{c}{0.270}\\
            \texttt{Linear SVM} & \multicolumn{1}{c}{0.622} & \multicolumn{1}{c}{0.064} & \multicolumn{1}{c}{0.117}\\
            \texttt{$k$-NN} & \multicolumn{1}{c}{0.517} & \multicolumn{1}{c}{0.398} & \multicolumn{1}{c}{0.450}\\
            \texttt{Kernel SVM} & \multicolumn{1}{c}{0.599} & \multicolumn{1}{c}{0.360} & \multicolumn{1}{c}{0.450}\\
            \texttt{Decision Tree} & \multicolumn{1}{c}{0.537} & \multicolumn{1}{c}{0.293} & \multicolumn{1}{c}{0.340}\\
            \texttt{STGP} ($\delta=10^{-5}$) & \multicolumn{1}{c}{0.457} & \multicolumn{1}{c}{0.968} & \multicolumn{1}{c}{0.621}\\
            \midrule[0.3pt]\bottomrule[1pt]
        \end{tabular}%
    \end{threeparttable}
}
\end{table}

We compare the hotspot detection results of our proposed method and several standard methods in binary classification, including perceptron, logistic regression, linear support vector machine (SVM), $k$-nearest neighbor ($k$-NN), kernel SVM with Gaussian kernel, and decision tree; see \cite{shalev2014understanding} for a detailed review of those machine learning algorithms and see Appendix~\ref{app:baseline} for hyperparameter choices. Table~\ref{tab:f1score} shows the $F_1$ score for the out-of-sample prediction at county-level using our method.
The result confirms that our model significantly outperforms other baseline methods. 


\section{Conclusion}
This paper proposes a Bayesian framework that combines hotspot detection and case prediction cohesively through a latent spatio-temporal random variable.
The latent variable is modeled by a Gaussian process, where a flexible non-stationary kernel function characterizes its covariance. 
The framework has shown immense promise in modeling and predicting the COVID-19 hotspots in the United States.
Our numerical study has also shown that the proposed kernel enjoys great representative power while being highly interpretable. 


%


\ifCLASSOPTIONcaptionsoff
  \newpage
\fi



\bibliographystyle{IEEEtran}
\bibliography{refs}
%


\clearpage
\newpage

\appendices

\section{Cross-validation for $\delta$}
\label{append:cv-exp-results}

\begin{figure}[h!]
\centering
\begin{subfigure}[h]{0.45\linewidth}
\includegraphics[width=\linewidth]{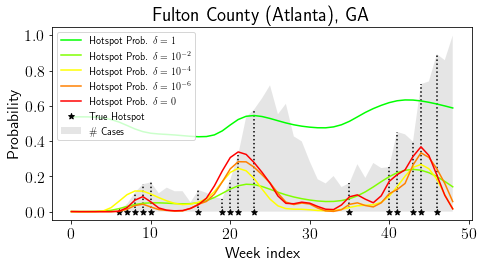}
\end{subfigure}
\begin{subfigure}[h]{0.45\linewidth}
\includegraphics[width=\linewidth]{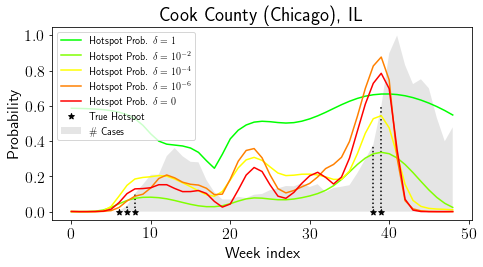}
\end{subfigure}
\vfill
\begin{subfigure}[h]{0.45\linewidth}
\includegraphics[width=\linewidth]{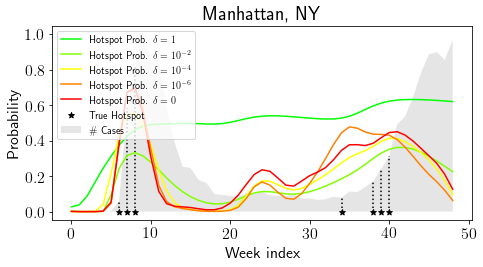}
\end{subfigure}
\begin{subfigure}[h]{0.45\linewidth}
\includegraphics[width=\linewidth]{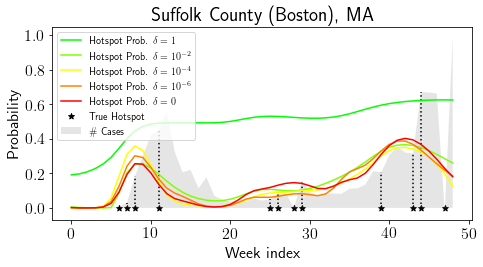}
\end{subfigure}
\vfill
\begin{subfigure}[h]{0.45\linewidth}
\includegraphics[width=\linewidth]{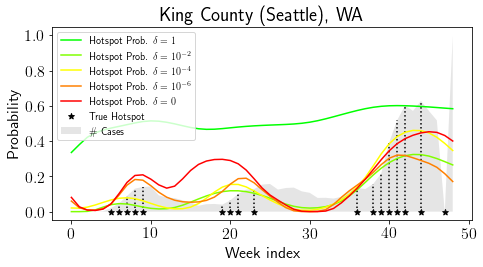}
\end{subfigure}
\begin{subfigure}[h]{0.45\linewidth}
\includegraphics[width=\linewidth]{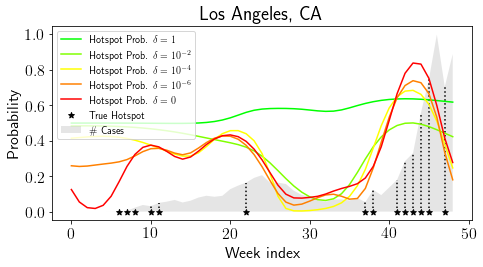}
\end{subfigure}
\vfill
\begin{subfigure}[h]{0.45\linewidth}
\includegraphics[width=\linewidth]{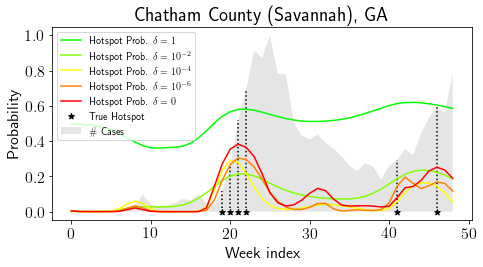}
\end{subfigure}
\begin{subfigure}[h]{0.45\linewidth}
\includegraphics[width=\linewidth]{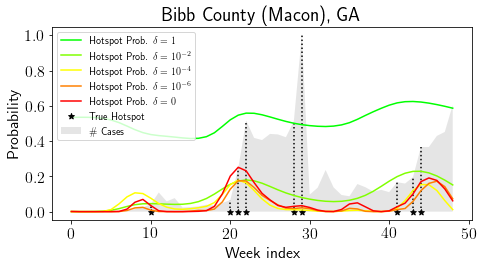}
\end{subfigure}
\caption{Comparison of one-week-ahead and county-wise hotspot probability $p(\mathbf{h}_*)$ using different $\delta$. The model with $\delta=10^{-5}$ attains the best performance in $F_1$ score.}
\label{fig:cv-delta-insample}
\end{figure}

\begin{figure}[h!]
\centering
\includegraphics[width=.8\linewidth]{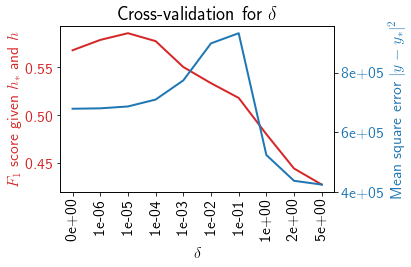}
\caption{Cross-validation result for $\delta$ displaying the mean-square error (blue) and $F_1$-score (red).}
\label{fig:cv-delta-outofsample}
\end{figure}

This section presents the cross-validation result for $\delta$ in \eqref{eq:objective} defined in Section~\ref{sec:spatio-temporal-gaussian}. Fig.~\ref{fig:cv-delta-insample} gives several examples of the predictions using different $\delta$. Fig.~\ref{fig:cv-delta-outofsample} summarizes the 5-fold cross-validation that quantitatively measures the $F_1$ score of the hotspot detection and the mean square error of the case prediction with different $\delta$. 

Note that the likelihood of observed confirmed cases plays a critical role in ``regularizing'' the \emph{hotspot model} and the optimal choice of $\delta$ ($\delta=10^{-5}$) leads to the best performance in hotspot prediction (out-of-sample). 
We also investigated the model with $\delta=0$ (an alternative model only including the hotspot component). 
The result shows that the $F_1$ score of out-of-sample prediction for the model with $\delta=0$ is 0.57, which is significantly lower than the model with $\delta=10^{-6}$.

\section{Proof of non-stationary kernel}
\label{append:proof-non-stationary-kernel}

Assume two independent bivariate Gaussian random variables $X_s$, $X_{s'}$ centered at locations $s, s'$, respectively, with $\Sigma_s, \Sigma_{s'}$ parameterized by
\[
\Sigma_s = \begin{bmatrix} a^2 & \rho a b \\ \rho a b & b^2 \end{bmatrix},~\Sigma_{s^{'}} = \begin{bmatrix} {a'}^2 & \rho' a' b' \\ \rho' a' b' & {b'}^2 \end{bmatrix}.
\]
Given two independent Gaussian random variables $X$ and $Y$ and their probability density functions $f_X$ and $f_Y$, the distribution $f_Z$ of $Z = X + Y$ equals the convolution of $f_X$ and $f_Y$, i.e., 
\[
    f_Z(z) = \int_{-\infty}^{\infty} f_Y(z-x) f_X(x) dx
\]
Denote the probability density function of $X_s$ and $X_{s'}$ as $\kappa_s(\cdot), \kappa_{s'}(\cdot)$, we have 
\[
f_{X_s+X_{s'}}(x) = \int_{\mathbb{R}^2} \kappa_s(u)\kappa_{s'}(x-u)du.
\] 
We also have the following equation due to the property of the Gaussian function:
\[
\kappa_{s'}(2s'-u) = \kappa_{s'}(u),~
\kappa_{s}(2s-u) = \kappa_s(u).
\]
Let $x = 2s'$ or $x = 2s$, we therefore have
\[
f_{X_s+X_{s'}}(2s') = f_{X_s+X_{s'}}(2s) = \int_{\mathbb{R}^2} \kappa_s(u)\kappa_{s'}(u)du = \upsilon(s, s'),
\]
which leads to \eqref{eq:def-non-stationary-kernel}.

Since $X_s + X_{s'}$ follows a Gaussian distribution
$
X_s + X_{s'} \sim \mathcal{N}\left ( s + s', \Sigma_s + \Sigma_{s'} \right )
$,
the non-stationary kernel $\upsilon(s, s')$ can be written as
\begin{align*}
    &~\upsilon(s, s') \\
    =&~f_{X_s+X_{s'}}(2s') \\
    =&~\frac{1}{2\pi | \Sigma_s + \Sigma_{s'} |^{\frac{1}{2}}} \exp\left \{-\frac{1}{2}(s' - s)^\top(\Sigma_s + \Sigma_{s'})^{-1}(s' - s)\right \}.
\end{align*}
Let $P = (\rho^2 - 1) b^2$ and $P' = ({\rho'}^2 - 1) {b'}^2$, we have 
\begin{align*}
    \upsilon(s, s') \propto \frac{1}{q_1} \exp\left\{-\frac{1}{q_2} (s-s')^\top W (s-s')\right\},
\end{align*}
where
\begin{align*}
    W = &~ \begin{bmatrix}
    b^2 + b'^2 & - (\rho a b + \rho' a' b')\\
    - (\rho a b + \rho' a' b') & a^2 + a'^2
    \end{bmatrix},\\
    q_1 = &~ 2 \pi\sqrt{-2\rho \rho' a a' b b' - a^2 (P - b'^2) - a'^2(P' - b^2)},\\
    q_2 = 
    &~ -2(2\rho \rho' a a' b b' + a^2 (P - b'^2) + a'^2(P' - b^2)).
\end{align*}

\section{Reparametrization of Gaussian distribution}
\label{append:reparametrization-gaussian}

Assume an ellipse centered at the origin with area $A$ has two focus points $(\psi_x, \psi_y), (-\psi_x, -\psi_y)$ in $\mathbb{R}^2$ where $\psi_x, \psi_y \in \mathbb{R}$. 
We define the semi-major and semi-minor axis of the ellipse as $\sigma_1, \sigma_2$. 
According to the ellipse formula we have:
\[
    \left \{ 
    \begin{array}{rl}
        \pi \sigma_1 \sigma_2 &= A, \\
        \sigma_1^2 - \sigma_2^2 &= \psi_x^2 + \psi_y^2 = \|\psi\|^2.
    \end{array}
    \right .
    \label{Ellipse}
\]
By solving the above linear equation system, we have
\begin{align*}
\sigma_1 = &~(\frac{\sqrt{4A^2 + \|\psi\|^4\pi^2}}{2\pi} + \frac{\|\psi\|^2}{2})^{\frac{1}{2}}, \\
\sigma_2 = &~(\frac{\sqrt{4A^2 + \|\psi\|^4\pi^2}}{2\pi} - \frac{\|\psi\|^2}{2})^{\frac{1}{2}}.
\end{align*}
Since the rotation angle $\alpha$ of the ellipse is $\alpha = \tan^{-1}(\psi_y/\psi_x)$, a bivariate normal random variable $X$ can be defined as
\[
X = 
\begin{bmatrix} 
\cos{\alpha} & -\sin{\alpha} \\ 
\sin{\alpha} & \cos{\alpha} 
\end{bmatrix} 
\begin{bmatrix}
Z_1 \\
Z_2
\end{bmatrix},
\]
where $Z_1$ and $Z_2$ are two independent random variables with variance $\sigma_1^2$ and $\sigma_2^2$, respectively.  
Here we introduce a kernel scale parameter $\tau_z$ and derive the covariance of $X$ as follows:
\begin{align*}
    \Sigma 
    = &~\tau_z^2 
    \begin{bmatrix} 
    \sigma_1^2\cos^2 \alpha + \sigma_2^2 \sin^2\alpha & (\sigma_1^2 - \sigma_2^2)\cos{\alpha}\sin{\alpha} \\
    (\sigma_1^2 - \sigma_2^2)\cos{\alpha}\sin{\alpha} & \sigma_1^2\sin^2 \alpha + \sigma_2^2 \cos^2\alpha 
    \end{bmatrix}
\end{align*}
Substitute the solution of $\sigma_1$ and $\sigma_2$ into the above equation, we have
\[
\begin{aligned}
  &~\sigma_1^2\cos^2 \alpha + \sigma_2^2 \sin^2\alpha \\
  = &~\frac{\sqrt{4A^2 + \|\psi\|^4\pi^2}}{2\pi} (\cos^2 \alpha + \sin^2 \alpha) + \frac{\|\psi\|^2}{2} (\cos^2 \alpha - \sin^2 \alpha)  \\
  = &~\frac{\sqrt{4A^2 + \|\psi\|^4\pi^2}}{2\pi} + \frac{\|\psi\|^2}{2} \cos(2\alpha),
\end{aligned}
\]
and similarly
\begin{align*}
    \sigma_1^2\sin^2 \alpha + \sigma_2^2 \cos^2\alpha = &~ \frac{\sqrt{4A^2 + \|\psi\|^4\pi^2}}{2\pi} - \frac{\|\psi\|^2}{2} \cos(2\alpha),\\
    (\sigma_1^2 - \sigma_2^2)\cos{\alpha}\sin{\alpha}=&~\|\psi\|^2\cos{\alpha}\sin{\alpha}=\frac{\|\psi\|^2}{2}\sin{2\alpha}.
\end{align*}
Thereby we obtain the matrix shown in Equation \eqref{eq:ellipse-covariance}.

\section{Derivation of ELBO}
\label{append:elbo}

Assume the posterior distribution  $p(\mathbf{f}, \mathbf{u}|\mathbf{y})$ over random variable vector $\mathbf{f}$ and $\mathbf{u}$ is approximated by a variational distribution $q(\mathbf{f}, \mathbf{u})$. 
Suppose this variational distribution $q(\mathbf{f}, \mathbf{u})$ can be factorized as $q(\mathbf{f}, \mathbf{u}) = p(\mathbf{f} | \mathbf{u}) q(\mathbf{u})$. Hence, the ELBO can be derived as follows:
\begingroup
\allowdisplaybreaks
\begin{align*}
    \log p(\mathbf{y})
    =&~ \log \int \int p(\mathbf{y}|\mathbf{f}, \mathbf{u}) p(\mathbf{f}, \mathbf{u}) d \mathbf{f} d \mathbf{u}\\
    =&~ \log \int \int p(\mathbf{y}|\mathbf{f}) p(\mathbf{f}, \mathbf{u}) d \mathbf{f} d \mathbf{u}\\
    =&~ \log \int \int p(\mathbf{y}|\mathbf{f}) p(\mathbf{f}, \mathbf{u}) \frac{q(\mathbf{f}, \mathbf{u})}{q(\mathbf{f}, \mathbf{u})} d \mathbf{f} d \mathbf{u}\\
    =&~ \log \mathbb{E}_{q(\mathbf{f}, \mathbf{u})} \left [  p(\mathbf{y}|\mathbf{f}) \frac{p(\mathbf{f}, \mathbf{u})}{q(\mathbf{f}, \mathbf{u})} \right ]\\
    \overset{(i)}{\ge} &~ \mathbb{E}_{q(\mathbf{f}, \mathbf{u})} \log \left [  p(\mathbf{y}|\mathbf{f}) \frac{p(\mathbf{f}, \mathbf{u})}{q(\mathbf{f}, \mathbf{u})} \right ]\\
    =&~ \int \int \log p(\mathbf{y}|\mathbf{f}) q(\mathbf{f}, \mathbf{u}) d \mathbf{f} d \mathbf{u} - \\
    &~\int \int \log \left ( \frac{q(\mathbf{f}, \mathbf{u})}{p(\mathbf{f}, \mathbf{u})} \right ) q(\mathbf{f}, \mathbf{u}) d \mathbf{f} d \mathbf{u}\\
    \overset{(ii)}{=}&~ \mathbb{E}_{q(\mathbf{f})}\left [ \log p(\mathbf{y}|\mathbf{f}) \right ] - \text{KL}\left [ q(\mathbf{u}) || p(\mathbf{u}) \right ],
\end{align*}
\endgroup
where $q(\mathbf{f})$ is the marginal of $\mathbf{f}$ from the joint variational distribution $q(\mathbf{f}, \mathbf{u})$, by integrating $\mathbf{u}$ out. The inequality $(i)$ holds due to the the Jensen's inequality. The equality $(ii)$ holds because
\begingroup
\allowdisplaybreaks
\begin{align*}
&~ \int \int \log \left ( \frac{q(\mathbf{f}, \mathbf{u})}{p(\mathbf{f}, \mathbf{u})} \right ) q(\mathbf{f}, \mathbf{u}) d \mathbf{f} d \mathbf{u}\\
= &~ \int \int \log \left ( \frac{p(\mathbf{f}| \mathbf{u}) q(\mathbf{u})}{p(\mathbf{f} |\mathbf{u}) p(\mathbf{u})} \right ) q(\mathbf{f}, \mathbf{u}) d \mathbf{f} d \mathbf{u}\\
= &~ \int \int \log \left (\frac{q(\mathbf{u})}{p(\mathbf{u})} \right ) q(\mathbf{f}, \mathbf{u}) d \mathbf{f} d \mathbf{u}\\
= &~ \int \log \left (\frac{q(\mathbf{u})}{p(\mathbf{u})} \right ) \left ( \int q(\mathbf{f}, \mathbf{u}) d \mathbf{f} \right) d \mathbf{u}\\
= &~ \int \log \left (\frac{q(\mathbf{u})}{p(\mathbf{u})} \right ) q(\mathbf{u}) d \mathbf{u}\\
= &~ \text{KL}\left [ q(\mathbf{u}) || p(\mathbf{u}) \right ].
\end{align*}
\endgroup

To calculate the ELBO, we also need to derive the analytical expression for $q(\mathbf{f})$ and $\text{KL}[q(\mathbf{u})||p(\mathbf{f})]$. 
First, given the joint distribution defined in \eqref{eq:joint-dist-f-u}, we apply the multivariate Guassian conditional rule and have the closed-form expression for the conditional distribution:
\[
    p(\mathbf{f}|\mathbf{u}) = \mathcal{N}(\mathbf{A} \mathbf{u}, \mathbf{B}),
\]
where $\mathbf{A} = \mathbf{K}_{XZ} \mathbf{K}_{ZZ}^{-1}$ and $\mathbf{B} = \mathbf{K}_{XX} - \mathbf{K}_{XZ} \mathbf{K}_{ZZ}^{-1} \mathbf{K}_{XZ}^\top$.
Now, due to the factorization $q(\mathbf{f}, \mathbf{u}) = p(\mathbf{f} | \mathbf{u}) q(\mathbf{u})$, we have
\begingroup
\allowdisplaybreaks
\begin{align*}
    q(\mathbf{f})
    =&~\int q(\mathbf{f}, \mathbf{u}) d\mathbf{u}\\
    =&~\int p(\mathbf{f} | \mathbf{u}) q(\mathbf{u}) d\mathbf{u}\\
    =&~\int \mathcal{N}(\mathbf{f}|\mathbf{A}\mathbf{u}, \mathbf{B})\cdot \mathcal{N}(\mathbf{u}|\mathbf{m},\mathbf{S}) d\mathbf{u}\\
    =&~\int \mathcal{N}(\mathbf{f}|\mathbf{A}\mathbf{m}, \mathbf{A}\mathbf{S}\mathbf{A}^\top + \mathbf{B})\cdot \mathcal{N}(\mathbf{u}|\mathbf{m},\mathbf{S}) d\mathbf{u}\\
    =&~\mathcal{N}(\mathbf{f}|\mathbf{A}\mathbf{m}, \mathbf{A}\mathbf{S}\mathbf{A}^\top + \mathbf{B}) \cdot \int \mathcal{N}(\mathbf{u}|\mathbf{m},\mathbf{S}) d\mathbf{u}\\
    =&~\mathcal{N}(\mathbf{f}|\mathbf{A}\mathbf{m}, \mathbf{K}_{XX} + \mathbf{A} (\mathbf{S} - \mathbf{K}_{ZZ}) \mathbf{A}^\top).
\end{align*}
\endgroup
Next, we derive the analytical expression for the Kullback–Leibler (KL) divergence in the ELBO, 
since both $q(\mathbf{u}) = \mathcal{N}(\mathbf{m}, \mathbf{S})$ and $p(\mathbf{u}) = \mathcal{N}(\mathbf{0},\mathbf{K}_{ZZ})$ are multivariate Gaussian distributions. 
Therefore, the KL divergence between $q(\mathbf{u})$ and $p(\mathbf{u})$ is:
\begin{align*}
    &~ \text{KL}\left [ q(\mathbf{u}) || p(\mathbf{u}) \right ] \\
    = &~ \frac{1}{2}\left( \log\left( \frac{\det(\mathbf{K}_{ZZ})}{\det(\mathbf{S})} \right) - M + \text{tr}(\mathbf{K}_{ZZ}^{-1} \mathbf{S}) + (\mathbf{0} - \mathbf{m})^\top \mathbf{K}_{ZZ}^{-1} (\mathbf{0} - \mathbf{m})\right),
\end{align*}
where $\det(\cdot)$ is the matrix determinant and $\text{tr}(\cdot)$ is the trace of matrix. 

\section{Derivation of predictive posterior}
\label{append:pred-posterior}

A Bayesian model makes predictions based on the posterior distribution. Given testing locations $\mathbf{X}_*$, we can derive the predictive posterior distribution $p(\mathbf{f}_*|\mathbf{y}, \mathbf{h})$:
\begingroup
\allowdisplaybreaks
\begin{align*}
    p(\mathbf{f}_*|\mathbf{y}) 
    = &~ \int \int p(\mathbf{f}_*, \mathbf{f}, \mathbf{u}|\mathbf{y}, \mathbf{h}) d \mathbf{f} d \mathbf{u}\\
    = &~ \int \int p(\mathbf{f}_* |\mathbf{f}, \mathbf{u}, \mathbf{y}, \mathbf{h}) p(\mathbf{f}, \mathbf{u} | \mathbf{y}, \mathbf{h}) d \mathbf{f} d \mathbf{u}\\
    = &~ \int \int p(\mathbf{f}_* |\mathbf{f}, \mathbf{u}) p(\mathbf{f}, \mathbf{u} | \mathbf{y}, \mathbf{h}) d \mathbf{f} d \mathbf{u}\\
    = &~ \int \int p(\mathbf{f}_* |\mathbf{f}, \mathbf{u}) q(\mathbf{f}, \mathbf{u}) d \mathbf{f} d \mathbf{u}\\
    = &~ \int \int p(\mathbf{f}_* |\mathbf{f}, \mathbf{u}) p(\mathbf{f} | \mathbf{u}) q(\mathbf{u}) d \mathbf{f} d \mathbf{u}\\
    = &~ \int \left ( \int p(\mathbf{f}_* |\mathbf{f}, \mathbf{u}) p(\mathbf{f} | \mathbf{u}) d \mathbf{f} \right ) q(\mathbf{u}) d \mathbf{u}\\
    = &~ \int p(\mathbf{f}_* | \mathbf{u}) q(\mathbf{u}) d \mathbf{u}.
\end{align*}
\endgroup

Similar to \eqref{eq:joint-dist-f-u}, since we assume that the unobserved future data comes from the same generation process, i.e., 
\[
    p([\mathbf{f}_*, \mathbf{u}]^\top) = \mathcal{N}\left(
    \mathbf{0},
    \begin{bmatrix}
    \mathbf{K}_{**} & \mathbf{K}_{*Z} \\
    \mathbf{K}_{*Z}^\top & \mathbf{K}_{*Z}
    \end{bmatrix}
    \right),
\]
we can apply the multivariate Gaussian conditional rule on the prior $p(\mathbf{f}_*, \mathbf{u})$ and obtain:
\begin{align*}
    p(\mathbf{f}_* | \mathbf{u})
    = &~\mathcal{N}(\mathbf{A}_* \mathbf{u}, \mathbf{B}_*),
\end{align*}
combining with $q(\mathbf{u}) = \mathcal{N}(\mathbf{m}, \mathbf{S})$, we have
\[
\int p(\mathbf{f}_* | \mathbf{u}) q(\mathbf{u}) d \mathbf{u} = \mathcal{N}(\mathbf{A}_* \mathbf{m}, \mathbf{A}_* \mathbf{S} \mathbf{A}_*^\top + \mathbf{B}_*)
\]
where $\mathbf{A}_* = \mathbf{K}_{*Z}\mathbf{K}_{ZZ}^{-1}$ and $\mathbf{B}_* = \mathbf{K}_{**} - \mathbf{K}_{*Z} \mathbf{K}_{ZZ}^{-1} \mathbf{K}_{*Z}^\top$.


\section{Model comparison with different $R$ in the spatial kernel}
\label{append:comp-r-exp-results}

\begin{figure}[h!]
\centering
\begin{subfigure}[h]{0.45\linewidth}
\includegraphics[width=\linewidth, frame]{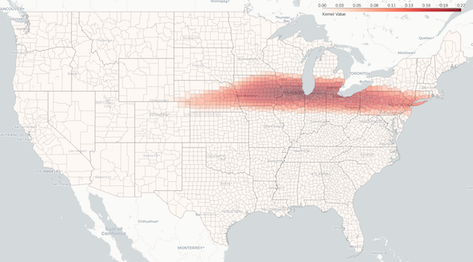}
\caption{$R=1$}
\end{subfigure}
\begin{subfigure}[h]{0.45\linewidth}
\includegraphics[width=\linewidth, frame]{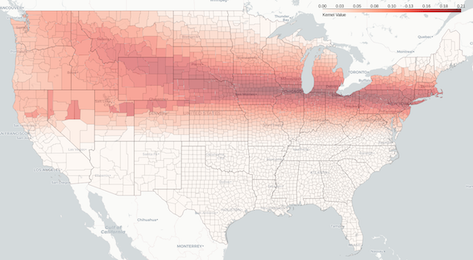}
\caption{$R=2$}
\end{subfigure}
\vfill
\begin{subfigure}[h]{0.45\linewidth}
\includegraphics[width=\linewidth, frame]{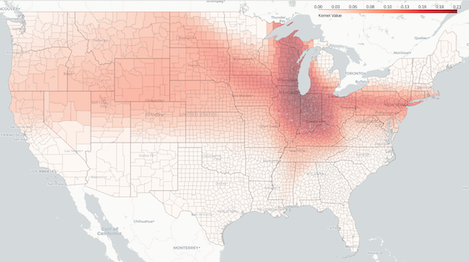}
\caption{$R=4$}
\end{subfigure}
\begin{subfigure}[h]{0.45\linewidth}
\includegraphics[width=\linewidth, frame]{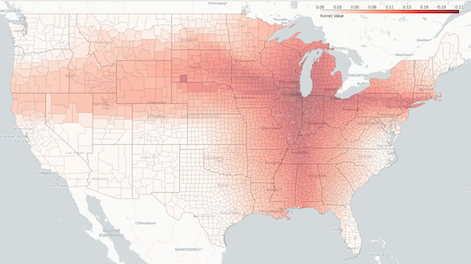}
\caption{$R=6$}
\end{subfigure}
\caption{visualization of spatial kernel at Chicago for different $R$.}
\label{fig:cv-delta}
\end{figure}

This section presents the comparison of our model using different number of spatial components in the kernel function. Fig.~\ref{fig:cv-delta} gives an example of visualized kernel evaluation centered at Chicago. It shows that the representative power of the kernel can be greatly enhanced by increasing the number of spatial components $R$.

\section{Cross-validation for neural network architecture}
\label{append:cv-nn-res}

\begin{figure}[!h]
    \centering
    \includegraphics[width=.8\linewidth]{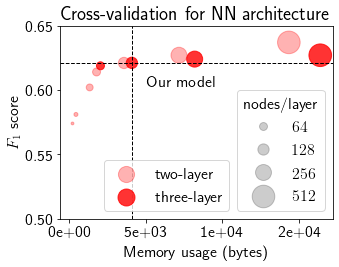}
    \caption{Cross-validation result for neural network structures. We use a three-layer neural network with 64 nodes per layer for each spatial kernel component, which achieves the balance between predictive performance and memory usage in practice.}
    \label{fig:cv-nn-outofsample}
\end{figure}

We choose the neural network architecture for each spatial kernel component by 5-fold cross-validation. 
Specifically, we compare the $F_1$ score of hotspot detection and its corresponding memory usage for 11 different neural network architectures under the same experimental setting. 
As shown in Fig.~\ref{fig:cv-nn-outofsample}, our model attains better predictive performance as the number of nodes or layers of the neural network grows. However, for any architecture with a number of nodes greater than 64, the performance gain starts to vanish while the GPU memory usage is increasing dramatically.
Therefore, we use a three-layer neural network with 64 nodes per layer for each spatial kernel component, which achieves a balance between predictive performance ($F_1$ score is 0.621) and memory usage (4,096 bytes per component) in practice.

\section{Detailed description for the baseline methods}
\label{app:baseline}

In this section, we provide a detailed description of the baseline methods used in Section~\ref{sec:compare_baseline_detect}, including the specific choice of hyperparameters. 

The perceptron classifier is an algorithm used for supervised learning, and its main component is a single-layer neural network. The perceptron classifier takes all the input feature values and computes their weighted sum. The weighted sum is then applied to the sign activation function, which outputs $1$ if the weighted sum is greater than zero and outputs $-1$ otherwise. The logistic regression is a similar method that takes all the input feature values and applies their weighted sum to the sigmoid function, which outputs the probability for the input feature to be in class $1$. The weights can be solved by maximum likelihood estimation through gradient descent. 

The linear support vector machine (SVM) is another type of binary classifier that aims to find the optimal linear decision boundary (hyperplane) to separate data from two classes in such a way that the separation is as wide as possible. The kernel SVM is a variant of linear SVM, and it considers non-linear separations; we use the Gaussian kernel with bandwidth parameter chosen as 0.1. 

The $k$-nearest neighbor ($k$-NN) classifier takes a test sample as input and uses the vote from its $k$-nearest neighbors as the output class; we set the number of neighbors to be $k=5$ and use the Euclidean distance to quantify pairwise distances between two samples features. 
Finally, the decision tree is a white box type of machine learning algorithm, and it has a flowchart-type tree structure. Each internal node represents the feature value, the branch represents the corresponding decision rule, and each leaf node represents the outcome. We set the maximum depth of the tree as four.

\end{document}